%% file: paper.tex
\documentclass[AMS,STIX1COL]{WileyNJD-v2}

\articletype{Research Article}

\received{17 September 2018}

\usepackage{amsmath}
\usepackage{hyperref}
\usepackage{graphics}
\usepackage{subfig}
\usepackage{url}

\usepackage{tikz, subfig, amsthm, multirow, float, enumerate}
\usepackage{tikz-3dplot}
\usetikzlibrary{arrows, shapes, positioning}
\usepackage{bm}

\newcommand{\nv}{{n_{\scriptscriptstyle V}}}
\newcommand{\nh}{{n_{\scriptscriptstyle H}}}

\providecommand{\tightlist}{%
  \setlength{\itemsep}{0pt}\setlength{\parskip}{0pt}}

\AtBeginDocument{
  \label{FirstPage2}
  
}

\begin{document}

\title{Properties and Bayesian fitting of restricted Boltzmann machines}
\authormark{KAPLAN \textsc{et al}}

\author[1]{Andee Kaplan*}
\author[2]{Daniel Nordman}
\author[2,3]{Stephen Vardeman}

\address[1]{
\orgdiv{Department of Statistical Science}, \orgname{Duke University}, \orgaddress{\state{North Carolina}, \country{USA}}}
\address[2]{
\orgdiv{Department of Statistics}, \orgname{Iowa State University}, \orgaddress{\state{Iowa}, \country{USA}}}
\address[3]{
\orgdiv{Department of Industrial and Manufacturing Systems Engineering}, \orgname{Iowa State University}, \orgaddress{\state{Iowa}, \country{USA}}}

\corres{*Andee Kaplan, Department of Statistical Science, Duke University, P.O. Box 90251,
Durham, NC 27708-0251. \email{\href{mailto:andrea.kaplan@duke.edu}{\nolinkurl{andrea.kaplan@duke.edu}}}}

\abstract[Summary]{
A restricted Boltzmann machine (RBM) is an undirected graphical model
constructed for discrete or continuous random variables, with two
layers, one hidden and one visible, and no conditional dependency within
a layer. In recent years, RBMs have risen to prominence due to their
connection to deep learning. By treating a hidden layer of one RBM as
the visible layer in a second RBM, a deep architecture can be created.
RBMs are thought to thereby have the ability to encode very complex and
rich structures in data, making them attractive for supervised learning.
However, the generative behavior of RBMs is largely unexplored and
typical fitting methodology does not easily allow for uncertainty
quantification in addition to point estimates. In this paper, we discuss
the relationship between RBM parameter specification in the binary case
and model properties such as degeneracy, instability and
uninterpretability. We also describe the associated difficulties that
can arise with likelihood-based inference and further discuss the
potential Bayes fitting of such (highly flexible) models, especially as
Gibbs sampling (quasi-Bayes) methods are often advocated for the RBM
model structure.
}

\maketitle

\hypertarget{introduction}{%
\section{Introduction}\label{introduction}}

The data mining and machine learning communities have recently shown
great interest in deep learning, specifically in stacked restricted
Boltzmann machines (RBMs) (see Salakhutdinov and Hinton 2009, 2012;
Srivastava, Salakhutdinov, and Hinton 2013; Le Roux and Bengio 2008 for
examples). A RBM is a probabilistic undirected graphical model (for
discrete or continuous random variables) with two layers, one hidden and
one visible, with no conditional dependency within a layer (Smolensky
1986). These models have reportedly been used with success in
classification of images (Larochelle and Bengio 2008; Srivastava and
Salakhutdinov 2012). However, the model properties are largely
unexplored in the literature and the commonly cited fitting methodology
remains heuristic-based and relies on rough approximation (Hinton,
Osindero, and Teh 2006). Additionally, the current fitting methodology
does not easily allow for interval estimation of parameters in the
model, and instead only provides for point estimation. In this paper, we
provide steps toward a fuller understanding of the model class and its
properties from the perspective of statistical model theory, and we then
explore the possibility of a rigorous fitting methodology with natural
uncertainty quantification. We find the RBM model class to be concerning
in two fundamental ways.

First, the models can be unsatisfactory as conceptualizations of how
data are generated. That is, recalling Fisher (1922), the aim of a
statistical model is to represent data in a compact way. Neyman and Box
further state that a model should ``provide an explanation of the
mechanism underlying the observed phenomena'' (Lehmann 1990; G. E. P.
Box 1967). At issue, simulation from RBMs can often produce data lacking
realistic variability so that such models may thereby fail to
satisfactorily reflect an observed data generation process. Such
behavior relates to model degeneracy (or near degeneracy), which is a
statistical concern in that many data processes of interest are
realistically not degenerate in their spectrum of potential outcomes.
For example, when sampling data from a nearly degenerate RBM used to
model an imaging process, only a small set of output possibilities
receives nearly all probability, and thus a sample of images will all be
copies of the same one image (or small number of images). An example of
ten 4-pixel images simulated from a nearly degenerate RBM model is
compared to ten 4-pixel images simulated from a non-degenerate RBM model
in Figure \ref{fig:sample-models}. The degenerate model places almost
all probability on one outcome, causing the image to be generated
repeatedly, whereas the non-degenerate model shows more potentially
realistic variation.

\par

\begin{figure}
\includegraphics[width=1\linewidth]{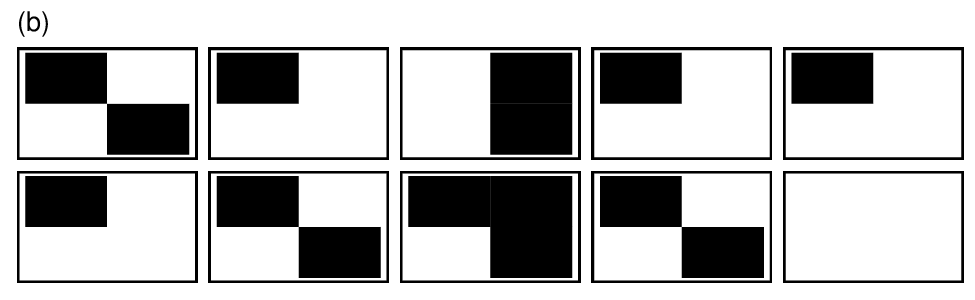} \includegraphics[width=1\linewidth]{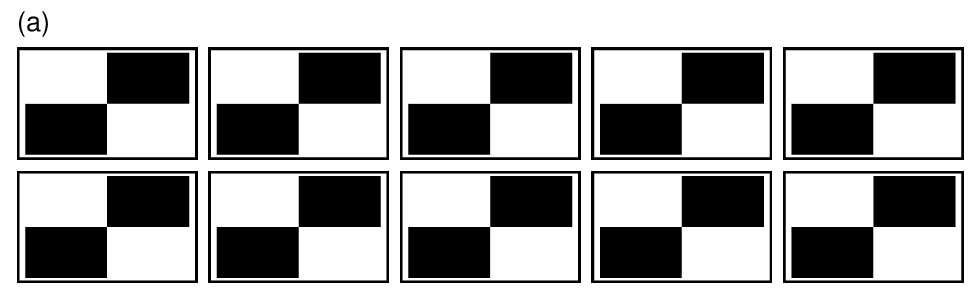} \caption{\label{fig:sample-models}Ten 4-pixel images simulated from a degenerate model (a) compared to ten 4-pixel images simulated from a non-degenerate model (b). The degenerate model places almost all probability on one outcome, causing the image to be generated repeatedly, whereas the non-degenerate model shows more realistic variation.}\label{fig:sample-models}
\end{figure}

In addition to such degeneracy, we find that RBMs can easily exhibit
types of instability, related to how sensitive the model probabilities
can be to small difference in data outcomes. In practice, this may be
seen when a single pixel change in an image results in a wildly
different classification in an image classification problem. Occurrences
of such behavior have recently been documented in RBMs (Li 2014), as
well as other deep architectures (Szegedy et al. 2013; Nguyen, Yosinski,
and Clune 2014). We describe (related) potential issues of model
instability, degeneracy and uninterpretability for the RBM class (which
are related properties) and examine the presence of these in Section
\ref{explorations-of-model-properties-through-simulation} through
simulations of small, manageable examples.

Separate from the exploration of model properties, we also investigate
the quality of estimation possible in fitting these models and examine
an estimation procedure that allows for uncertainty quantification
rather than point estimates alone. The fitting of RBMs can be
problematic for two reasons, the first being computational and the
second concerning flexibility. As the size of these models grows, both
maximum likelihood and Bayesian methods of fitting quickly become
intractable. The literature often suggests Markov chain Monte Carlo
(MCMC) tools for approximate maximization of likelihoods to fit these
models (e.g., Gibbs sampling to exploit conditional structure in hidden
and visible variables), but little is said about the attributes of
realized estimates (Hinton 2010; Hinton, Osindero, and Teh 2006),
including the absence of interval estimates.

Furthermore, these MCMC algorithms involve updating potentially many
latent variables (hiddens) which can critically influence convergence in
MCMC-based likelihood methods. Applying basic statistical principles in
fitting RBM models of tractable size, we compare three fully Bayesian
techniques involving MCMC, which are computationally more accessible
than direct maximum likelihood and also aim to avoid parts of a RBM
parameter space that yield unattractive models. As might be expected,
with greater computational complexity comes an increase in fitting
accuracy, but at the cost of practical feasibility.

As a factor compounding the computational challenges, issues in model
fitting traceable to model flexibility are potentially more concerning.
For a RBM model with enough hidden variables, any distribution for the
visibles can be approximated arbitrarily well (Le Roux and Bengio 2008;
Montufar and Ay 2011; and Montúfar, Rauh, and Ay 2011). However, for
binary data, the empirical distribution of an observed training set of
visible variables provides a highest likelihood benchmark, before any
parametric model class is even introduced and applied to obtain a
refinement of model fit. As a consequence, we find that any fully
principled fitting method based on the likelihood for a RBM with enough
hidden variables will seek to reproduce the (discrete) empirical
distribution of a training set. This aspect can be undesirable, and
perhaps even unexpected compared to most modeling scenarios, in that no
``smoothed distribution'' may result when fitting a RBM model of
sufficient size with a rigorous likelihood-based method. We are
therefore led to be skeptical that models that involve these structures
(like deep Boltzmann machines) can achieve useful prediction or
inference in a principled way without intentionally limiting the
flexibility of the fitted model.

Notions of weight-decay (penalization) and sparsity (regularization)
have been suggested in the RBM literature as practical remedies for both
over-fitting and poor mixing in the Markov chain during fitting (Hinton
2010; Tieleman 2008; Cho, Ilin, and Raiko 2012). Both \(L_1\) and
\(L_2\) type penalties are mentioned to achieve weight-decay, though the
benefits of these particular forms are unknown in different situations.
The degree to which regularization and penalization are used by
practitioners is not clear because these concepts are not (by default) a
part of the standard overall model-specification-and-fitting methodology
(Hinton 2002; Carreira-Perpinan and Hinton 2005). In this paper, we
attempt to address the concerns with overfitting and poor mixing in an
alternative (and perhaps more transparent) manner, via specification of
a Bayesian prior in Section \ref{bayesian-model-fitting}.

This paper is structured as follows. Section \ref{background} formally
defines the RBM including the joint distribution of hidden and visible
variables and explains the model's connection to deep learning.
Additionally, measures of model impropriety and methods of
quantifying/detecting it are defined. Section
\ref{explorations-of-model-properties-through-simulation} details our
explorations into the model behavior and potential propriety issues with
the RBM class. This work aligns with the call to action by Google
scientists in Sculley et al. (2018) to employ small toy examples to
advance empirical understanding of deep learning models and strive for a
``higher level of empirical rigor in the field.'' We examine three
Bayesian fitting techniques intended to avoid model impropriety issues
raised in Section \ref{model-fitting} and conclude with a discussion in
Section \ref{discussion}. Supplementary online material provides proofs
for results on RBM parameterizations and data codings described in
Section \ref{data-encoding}.

While applications of the RBM have been claimed to produce some
successes in classification problems, it is unclear if the model class
allows one to go beyond fitting to other statistical matters of
importance in using the models, such as quantification of the
uncertainty of estimation and the formulation of predictive
distributions. These are closely tied to the probability properties of
the RBM model for explaining data generation and our exposition intends
to contribute to a better understanding in this direction.

\hypertarget{background}{%
\section{Background}\label{background}}

\hypertarget{restricted-boltzmann-machines}{%
\subsection{Restricted Boltzmann
machines}\label{restricted-boltzmann-machines}}

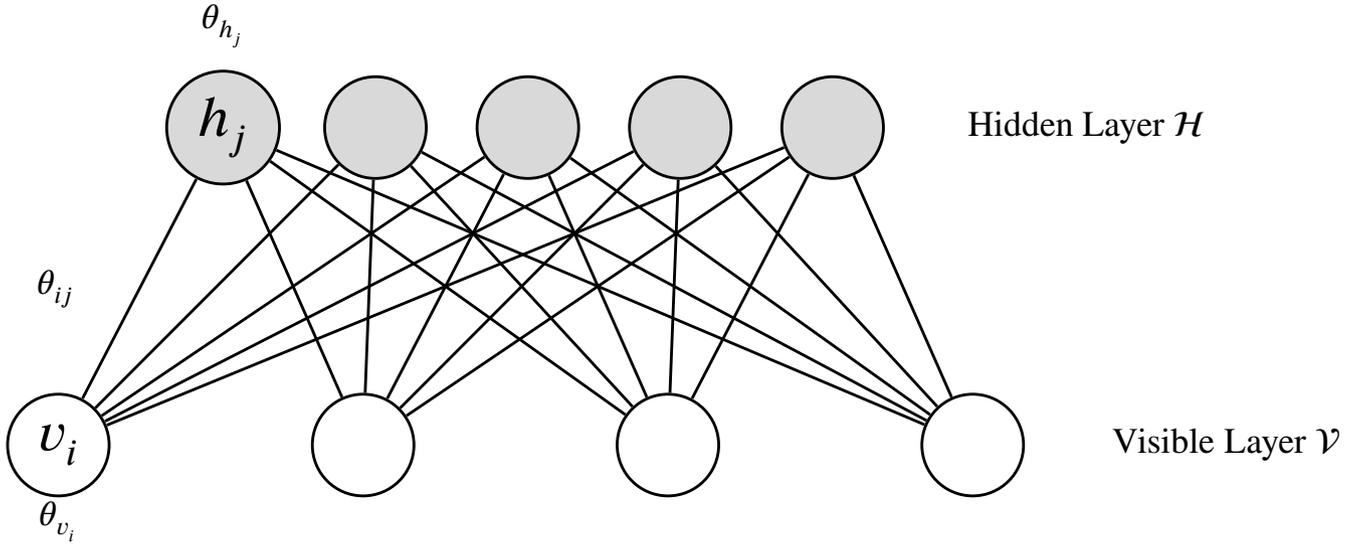
\begin{figure}
  \centering
  \resizebox{\linewidth}{!}{\input{rbm.tikz}}
  \caption{An example restricted Boltzmann machine (RBM), consisting of two layers, one hidden ($\mathcal{H}$) and one visible ($\mathcal{V}$), with no connections within a layer. Hidden nodes are indicated by gray filled circles and the visible nodes indicated by unfilled circles.}
  \label{fig:rbm}
\end{figure}

A restricted Boltzmann machine (RBM) is an undirected graphical model
specified for discrete or continuous random variables, binary variables
being most commonly considered. In this paper, we consider the binary
case for concreteness. A RBM architecture has two layers, hidden
(\(\mathcal{H}\)) and visible (\(\mathcal{V}\)), with no dependency
connections within a layer. An example of this structure is in Figure
\ref{fig:rbm} with the hidden nodes indicated by gray circles and the
visible nodes indicated by white circles.

A common use for RBMs is to create features for use in classification.
For example, binary images can be classified through a process that
treats the pixel values as the visible variables \(v_i\) in a RBM model
(Hinton, Osindero, and Teh 2006).

\hypertarget{joint-distribution}{%
\subsubsection{Joint distribution}\label{joint-distribution}}

Let \(\boldsymbol x = (h_1, \dots, h_\nh, v_1,\dots,v_\nv)\) represent
the states of the visible and hidden nodes in a RBM for some integers
\(\nv, \nh \ge 1\). Each single binary random variable, visible or
hidden, will take its values in a common coding set \(\mathcal{C}\),
where we allow (one of) two possibilities for the coding set,
\(\mathcal{C}=\{0,1\}\) or \(\mathcal{C}=\{-1,1\}\), with ``\(1\)''
always indicating the ``high'' value of the variable. While
\(\mathcal{C}=\{0,1\}\) may be a natural starting point, we argue in
Section \ref{explorations-of-model-properties-through-simulation} that
the coding \(\mathcal{C}=\{-1,1\}\) induces more interpretable model
properties for the RBM. A standard parametric form for probabilities
corresponding to a potential vector of states,
\(X = (H_1, \dots, H_\nh, V_1,\dots,V_\nv)\), for the nodes is
\begin{align}
\label{eq:pmf}
f_{\boldsymbol \theta} (\boldsymbol x)\equiv P_{\boldsymbol \theta}(\boldsymbol X = \boldsymbol x) = \frac{\exp\left(\sum\limits_{i = 1}^\nv \sum\limits_{j=1}^\nh \theta_{ij} v_i h_j + \sum\limits_{i = 1}^\nv\theta_{v_i} v_i + \sum\limits_{j = 1}^\nh\theta_{h_j} h_j\right)}{\gamma(\boldsymbol \theta)}, \quad \boldsymbol x \in \mathcal{C}^{\nh + \nv} 
\end{align} where
\(\boldsymbol \theta \equiv (\theta_{11}, \dots, \theta_{1\nh}, \dots, \theta_{\nv 1}, \dots, \theta_{\nv \nh}, \theta_{v_1}, \dots, \theta_{v_\nv}, \theta_{h_1}, \dots, \theta_{h_\nh}) \in \mathbb{R}^{\nv + \nh + \nv*\nh}\)
denotes the vector of model parameters and the denominator \[
\gamma(\boldsymbol \theta) = \sum_{\boldsymbol x \in \mathcal{C}^{\nh+\nv}}\exp\left(\sum_{i = 1}^\nv \sum_{j=1}^\nh \theta_{ij} v_i h_j + \sum_{i = 1}^\nv\theta_{v_i} v_i + \sum_{j = 1}^\nh\theta_{h_j} h_j\right)
\] is the normalizing function that ensures the probabilities
\eqref{eq:pmf} sum to one. For
\(\boldsymbol x = (h_1, \dots, h_\nh, v_1, \dots, v_\nh) \in \mathcal{C}^{\nv + \nh}\)
and \begin{align}
\boldsymbol t(\boldsymbol x) &= (h_1, \dots, h_\nh, v_1, \dots, v_\nv, v_1 h_1, \dots, v_\nv h_\nh) \in \mathcal{C}^{\nh + \nv + \nh*\nv}, \label{eq:t}
\end{align} let
\(\mathcal{T} = \{\boldsymbol t(\boldsymbol x): \boldsymbol x \in \mathcal{C}^{\nv + \nh}\} \subset \mathbb{R}^{\nv + \nh + \nv * \nh}\)
be the set of possible values for the vector of variables needed to
compute probabilities \eqref{eq:pmf} in the model, and write
\(Q_{\boldsymbol \theta}(\boldsymbol x) = \sum\limits_{i = 1}^\nh\sum\limits_{j=1}^\nv \theta_{ij} h_i v_j + \sum\limits_{i = 1}^\nh\theta_{hi} h_i + \sum\limits_{j = 1}^\nv\theta_{vj} v_j\)
for the ``neg-potential'' function. The RBM model is parameterized by
\(\boldsymbol \theta\) containing two types of parameters, main effects
and interaction effects. The main effects parameters
(\(\{\theta_{v_i}, \theta_{h_j}\}_{\substack{i = 1, \dots, \nv,\\j = 1, \dots, \nh}}\))
weight the values of the visible \(v_i\) and hidden \(h_j\) nodes in
probabilities \eqref{eq:pmf} and the interaction effect parameters
(\(\theta_{ij}\)) weight the values of the connections \(v_i h_j\), or
dependencies, between hidden and visible layers.

Due to the potential size of the model, the normalizing constant
\(\gamma(\boldsymbol \theta)\) can be practically impossible to
calculate, making simple estimation of the model parameter vector
problematic. In model fitting, a kind of Gibbs sampling can be tried due
to the conditional architecture of the RBM (i.e.~visibles given hiddens
or vice verse). Specifically, the conditional independence of nodes in
each layer (given those nodes in the other layer) allows for stepwise
simulation of both hidden layers and model parameters (e.g., see the
contrastive divergence of Hinton (2002) or Bayes methods in Section
\ref{model-fitting}).

\hypertarget{connection-to-deep-learning}{%
\subsubsection{Connection to Deep
Learning}\label{connection-to-deep-learning}}

RBMs have risen to prominence in recent years due to their connection to
deep learning (see Hinton, Osindero, and Teh 2006; Salakhutdinov and
Hinton 2012; Srivastava, Salakhutdinov, and Hinton 2013 for examples).
By stacking multiple layers of RBMs in a deep architecture, proponents
of the methodology claim to produce the ability to learn ``internal
representations that become increasingly complex, which is considered to
be a promising way of solving object and speech recognition problems''
(Salakhutdinov and Hinton 2009, 450). The stacking is achieved by
treating a hidden layer of one RBM as the visible layer in a second RBM,
and so on, until the desired multi-layer architecture is created.

\hypertarget{degeneracy-instability-and-uninterpretability}{%
\subsection{Degeneracy, instability, and
uninterpretability}\label{degeneracy-instability-and-uninterpretability}}

The highly flexible nature of a RBM (having as it does
\(\nh + \nv + \nh*\nv\) parameters) creates at least three kinds of
potential issues of model impropriety that we will call
\emph{degeneracy}, \emph{instability}, and \emph{uninterpretability}. In
this section we define these, consider how to quantify them in a RBM,
and point out relationships among them.

\hypertarget{near-degeneracy}{%
\subsubsection{Near-degeneracy}\label{near-degeneracy}}

In Random Graph Model theory, \emph{model degeneracy} means there is a
disproportionate amount of probability placed on only a few elements of
the sample space, \(\mathcal{X}\), by the model (Handcock 2003). For
random graph models, \(\mathcal{X}\) denotes all possible graphs that
can be constructed from a set of nodes and an exponentially
parameterized random graph model has a distribution with a probability
mass function of the form \begin{align*}
f_{\boldsymbol \theta} (\boldsymbol x) = \frac{\exp\left(\boldsymbol \theta^T \boldsymbol t(\boldsymbol x)\right)}{\gamma(\boldsymbol \theta)}, \boldsymbol x \in \mathcal{X},
\end{align*} where
\(\boldsymbol \theta \in \Theta \subset \mathbb{R}^q\) is the model
parameter, and \(\boldsymbol t: \mathcal{X} \rightarrow \mathbb{R}^q\)
is a vector of statistics based on the adjacency matrix of a graph.
Here, as earlier for RBMs,
\(\gamma(\boldsymbol \theta) = \sum_{\boldsymbol x \in \mathcal{X}} \exp\left(\boldsymbol \theta^T \boldsymbol t(\boldsymbol x)\right)\)
is the normalizing function. Let \(C\) denote the convex hull of the
potential outcomes of sufficient statistics,
\(\{\boldsymbol t(\boldsymbol x): \boldsymbol x \in \mathcal{X}\}\),
under the model above. Handcock (2003) classifies an exponentially
parameterized random graph model at \(\boldsymbol \theta\) as
\emph{near-degenerate} if the mean value of the vector of sufficient
statistics under \(\boldsymbol \theta\),
\(\boldsymbol \mu(\boldsymbol \theta) = \text{E}_{\boldsymbol \theta}\boldsymbol t( \boldsymbol X)\),
is close to the boundary of \(C\). Intuitively, if a model is
near-degenerate in the sense that only a small number of elements of the
sample space \(\mathcal{X}\) have positive probability, the expected
value \(\text{E}_{\boldsymbol \theta}\boldsymbol t( \boldsymbol X)\) is
an average of that same small number of values of \(t( \boldsymbol x)\)
(defining the boundary of the hull \(C\)) and can be expected to
\emph{not} be pulled deep into the interior of \(C\).

A RBM model can be thought to exhibit an analogous form of
\emph{near-degeneracy} when there is a disproportionate amount of
probability placed on a small number of elements in the sample space of
visibles and hiddens, \(\mathcal{C}^{\nv + \nh}\). Using the idea of
Handcock (2003), when the random vector
\(\boldsymbol t(\boldsymbol x) = (v_1, \dots, v_\nv, h_1, \dots, h_\nh, v_1 h_1, \dots, v_V h_\nh ) \in \mathcal{T} \equiv \{\boldsymbol t(\boldsymbol x): \boldsymbol x \in \mathcal{C}^{\nh + \nv} \}\)
from \eqref{eq:t}, appearing in the neg-potential function
\(Q_{\boldsymbol \theta}(\cdot)\), has a mean vector
\(\boldsymbol \mu(\boldsymbol \theta) \in \mathbb{R}^{\nv + \nh + \nv * \nh}\)
close to the boundary of the convex hull of \(\mathcal{T}\), and the RBM
model can be said to exhibit near-degeneracy at
\(\boldsymbol \theta \in \mathbb{R}^{\nv + \nh + \nh*\nv}\). Here the
mean of \(\boldsymbol t(\boldsymbol x)\) is \begin{align*}
\boldsymbol \mu(\boldsymbol \theta) = \text{E}_{\boldsymbol \theta} \boldsymbol t(\boldsymbol X)  &= \sum\limits_{x \in \mathcal{C}^{\nv + \nh}} \left\{ \boldsymbol t(\boldsymbol x) f_{\boldsymbol \theta} (\boldsymbol x) \right\} \\
&= \sum\limits_{x \in \mathcal{C}^{\nv + \nh}} \left\{ \boldsymbol t(\boldsymbol x)\dfrac{\exp\left(\sum\limits_{i = 1}^\nv \sum\limits_{j=1}^\nh \theta_{ij} v_i h_j + \sum\limits_{i = 1}^\nv\theta_{v_i} v_i + \sum\limits_{j = 1}^\nh\theta_{h_j} h_j\right)}{\sum\limits_{\boldsymbol x \in \mathcal{C}^{\nh + \nv}}\exp\left(\sum\limits_{i = 1}^\nv \sum\limits_{j=1}^\nh \theta_{ij} v_i h_j + \sum\limits_{i = 1}^\nv\theta_{v_i} v_i + \sum\limits_{j = 1}^\nh\theta_{h_j} h_j\right)}\right\}.
\end{align*}

\hypertarget{instability}{%
\subsubsection{Instability}\label{instability}}

Considering exponential families of distributions, Schweinberger (2011)
introduced a concept of model deficiency related to \emph{instability}.
Instability can be roughly thought of as excessive sensitivity in the
model, where small changes in the components of potential data outcomes,
\(\boldsymbol x\), may lead to substantial changes in the probability
mass function \(f_{\boldsymbol \theta}(\boldsymbol x)\). Furthermore,
model instability can be viewed on a spectrum of potential sensitivity
in probability structure, with model degeneracy included as a extreme or
limiting case of instability. To quantify ``instability'' more
rigorously (particularly beyond the definition given by Schweinberger
(2011)) it is useful to consider how RBM models might be expanded to
incorporate more and more visibles. When increasing the size of RBM
models, it becomes necessary to grow the number of model parameters (and
in this process one may also arbitrarily expand the number of hidden
variables used). To this end, let
\(\boldsymbol \theta_{\nv} \equiv (\theta_{v_1}, \dots, \theta_{v_\nv}, \theta_{h_1}, \dots, \theta_{h_\nh}, \theta_{11}, \dots, \theta_{\nv \nh}), \nv \ge 1\),
denote an element of a sequence of RBM parameters indexed by the number
\(\nv\) of visibles \((V_1, \dots, V_\nv)\) and define a log-ratio of
extremal probabilities (LREP) of the RBM model at
\(\boldsymbol \theta_{\nv}\) as \begin{align}
\frac{1}{\nv} \log \left[\frac{\max\limits_{(v_1, \dots, v_\nv) \in \mathcal{C}^\nv}P_{\boldsymbol \theta_\nv}(v_1, \dots, v_\nv)}{\min\limits_{(v_1, \dots, v_\nv) \in \mathcal{C}^\nv}P_{\boldsymbol \theta_\nv}(v_1, \dots, v_\nv)}\right] \equiv \frac{1}{\nv} \text{LREP}(\boldsymbol \theta_\nv) \label{eq:elpr}
\end{align} where
\(P_{\boldsymbol \theta_\nv}(v_1, \dots, v_\nv) \propto \sum_{(h_1, \dots, h_\nh) \in \mathcal{C}^\nh}\exp\left(\sum_{i = 1}^\nv \theta_{v_i} v_i + \sum_{j = 1}^\nh \theta_{h_j} h_j + \sum_{i = 1}^\nv \sum_{j = 1}^\nh \theta_{ij} v_i h_j\right)\)
is the RBM probability of observing outcome \((v_1, \dots, v_\nv)\) for
the visible variables \((V_1, \dots, V_\nv)\) under parameter vector
\(\boldsymbol \theta_\nv\), after marginalization of hidden variables.

In formulating a RBM model for a potentially large number of visibles
(i.e., as \(\nv \rightarrow \infty\)), we will say that the ratio
\eqref{eq:elpr} needs to stay bounded for a sequence of RBM models to be
stable. That is, we make the following convention.

\begin{definition}[S-unstable RBM]
Let $\boldsymbol \theta_\nv \in \mathbb{R}^{\nv + \nh +  \nh*\nv}, \nv \ge 1$, be an element of a sequence of RBM parameters where the number of hiddens, $\nh \equiv \nh(\nv) \ge 1$, can be an arbitrary function of the number $\nv$ of visibles. A RBM model formulation is \emph{Schweinberger-unstable} or \emph{S-unstable} if
\begin{align*}
\lim\limits_{\nv \rightarrow \infty} \frac{1}{\nv} \text{LREP}(\boldsymbol \theta_\nv) = \infty.
\end{align*}
\end{definition}

In other words, a RBM model sequence is unstable if, given any
\(c > 0\), there exists an integer \(n_c > 0\) so that
\(\frac{1}{\nv}\text{LREP}(\boldsymbol \theta_\nv) > c\) for all
\(\nv \ge n_c\). This definition of \emph{S-unstable} is a
generalization or re-interpretation of the ``unstable'' concept of
Schweinberger (2011) in that here RBM models for visibles
\((v_1, \dots, v_\nv)\) do not form an exponential family and the
dimensionality of \(\boldsymbol \theta_\nv\) is not fixed, but rather
grows with \(\nv\).

S-unstable RBM model sequences are undesirable for several reasons. One
is that, as mentioned above, small changes in data can lead to
overly-sensitive changes in probability under the data model. Consider,
for example, \[
\Delta(\boldsymbol \theta_\nv) \equiv \max \left\{\log \frac{P_{\boldsymbol \theta_\nv}(\boldsymbol v)}{P_{\boldsymbol \theta_\nv}(\boldsymbol v^*)} : \boldsymbol v \text{ }\& \text{ } \boldsymbol v^* \in \mathcal{C}^\nv \text{ differ in exactly one component}\right\},
\] denoting the biggest log-probability ratio for a one-component change
in data outcomes (visibles) at a RBM parameter
\(\boldsymbol \theta_\nv\). We then have the following result.

\begin{proposition}
\label{prp:instab1}
Let $c > 0$ and let $\text{LREP}(\boldsymbol \theta_\nv)$ be as in \eqref{eq:elpr} for an integer $\nv \ge 1$. If $\frac{1}{\nv}\text{LREP}(\boldsymbol \theta_\nv) > c$, then $\Delta(\boldsymbol \theta_\nv) > c$.
\end{proposition}

In other words, if the probability ratio \eqref{eq:elpr} is too large,
then a RBM model sequence will exhibit large probability shifts for very
small changes in data configurations (i.e., will exhibit instability).
Such instability can be a concern in a model for reasons similar to
those related to degeneracy: as outcomes vary over the sample space, the
geography of probabilities is extremely rugged, with deep pits following
sharp mountains. Recall the applied example of RBM models as a means to
classify images. For data as pixels in an image, the instability result
in Proposition \ref{prp:instab1} manifests itself as a one pixel change
in an image (one component of the visible vector) resulting in a large
shift in the probability, which in turn could result in a vastly
different classification of the image. Examples of this kind of behavior
have been presented in Szegedy et al. (2013) for deep learning models,
in which a one pixel change in a test image results in a wildly
different classification.

Additionally, S-unstable RBM model sequences may be formally connected
to the near-degeneracy of Section \ref{near-degeneracy} (in which model
sequences place all probability on a small portion of their sample
spaces). To see this, define an arbitrary modal set of possible outcomes
(i.e.~set of highest probability outcomes) in RBM models with parameters
\(\boldsymbol \theta_\nv, \nv \ge 1\) as \[
M_{\epsilon, \boldsymbol \theta_\nv} \equiv \left\{\boldsymbol v \in \mathcal{C}^\nv: \log P_{\boldsymbol \theta_\nv}(\boldsymbol v) > (1-\epsilon)\max\log\limits_{\boldsymbol v^*}P_{\boldsymbol \theta_\nv}(\boldsymbol v^*) + \epsilon\min\log\limits_{\boldsymbol v^*}P_{\boldsymbol \theta_\nv}(\boldsymbol v^*) \right\}
\] for a given \(0 < \epsilon < 1\). Then S-unstable model sequences are
guaranteed to be degenerate, as the following result shows.

\begin{proposition}
\label{prp:degen}
For an S-unstable RBM model sequence and any $0 < \epsilon < 1$, 
$$
P_{\boldsymbol \theta_\nv}\left((v_1, \dots, v_\nv) \in M_{\epsilon, \boldsymbol \theta_\nv}\right) \rightarrow 1 \text{ as } \nv \rightarrow \infty.
$$
\end{proposition}

In other words, S-unstable RBM model sequences are guaranteed to stack
up all probability on a specific set of outcomes for visibles, which
could potentially be arbitrarily narrow. Proofs of Propositions
\ref{prp:instab1} and \ref{prp:degen} follow from more general results
in Kaplan, Nordman, and Vardeman (2017). These findings also have
counterparts in results in Schweinberger (2011), but unlike results
there, are not limited in consideration to exponential family forms with
a fixed number of parameters.

\hypertarget{uninterpretability}{%
\subsubsection{Uninterpretability}\label{uninterpretability}}

For spatial Markov models, Kaiser (2007) defines a measure of model
impropriety he calls \emph{uninterpretability}, which is characterized
by dependence parameters in a model being so extreme that marginal
mean-structures fail to hold as anticipated by consideration of a model
statement. We adapt this notion to RBM models. Note that in a RBM, the
parameters \(\theta_{v_1}, \dots, \theta_{v_\nv}\) and
\(\theta_{h_1}, \dots, \theta_{h_\nh}\) are naturally associated with
main effects of visible and hidden variables and can be interpreted as
(logit functions of) means for variables
\(V_1, \dots, V_\nv, H_1, \dots, H_\nh\) in a model with no interaction
parameters, \(\theta_{ij} = 0, i = 1, \dots, \nv, j = 1, \dots, \nh\).
That is, with no interaction parameters, we have from \eqref{eq:pmf}
that \[
P_\theta(V_i=1) \propto e^{\theta_{v_i}} \quad \text{ and } \quad P_\theta(H_j=1) \propto e^{\theta_{h_j}}, \quad  i=1,\ldots,\nv,j=1,\ldots,\nh
\] so that, for example, \(\text{logit}(P_\theta(V_i=1))=\theta_{v_i}\)
(or \(2 \theta_{v_i}\)) under the coding \(\mathcal{C}=\{0,1\}\) (or
\(\{-1,1\}\)). Hence, these main effect parameters have a clear
interpretation under an independence model (one with
\(\theta_{ij} = 0\)) but this interpretation can break down as
interaction parameters increase in magnitude relative to the size of the
main effects. In such cases, the main effect parameters \(\theta_{v_1}\)
and \(\theta_{h_j}\) are no longer interpretable in the models
(statements of marginal means) and the dependence parameters are so
large as to dominate the entire model probability structure (also
destroying simple interpretation of dependence parameters
\(\theta_{ij}\), \(j=1,\ldots,\nh\) as local conditional modifications
of an overall marginal mean structure \(\theta_{v_i}\), as appearing for
example in the (logit) conditional probability
\(\mathrm{logit P_\theta(V_i=1|H_1,\ldots,H_{\nh})}= \theta_{v_i} + \sum_{j=1}^{\nh} \theta_{ij} H_j\)).
Whether or not parameter interpretation is itself a goal in a given
application of RBM models, this concept of interpretation can provide an
additional device for examining other aspects of model propriety related
to instability and degeneracy. As explained by Kaiser (2007), models
with interpretable dependence parameters typically correspond to
non-degenerate models, while degradation in interpretability is often
associated with model drift into degeneracy. To assess which parameter
values \(\boldsymbol \theta\) may cause difficulties in interpretation,
we use the difference
\(\text{E}\left[\boldsymbol X | \boldsymbol \theta\right] - \text{E}\left[\boldsymbol X | \boldsymbol \theta^* \right]\)
between two model expectations:
E\(\left[\boldsymbol X | \boldsymbol \theta\right]\) at
\(\boldsymbol \theta\) and expectations
E\(\left[\boldsymbol X | \boldsymbol \theta^* \right]\) where
\(\boldsymbol \theta^*\) matches \(\boldsymbol \theta\) for all main
effects but otherwise has \(\theta_{ij} = 0\) for
\(i = 1, \dots, \nv, j = 1, \dots, \nh\). (Hence,
\(\boldsymbol \theta^*\) and \(\boldsymbol \theta\) have the same main
effects but \(\boldsymbol \theta^*\) has \(0\) dependence parameters.)
Uninterpretability is then avoided at a parametric specification
\(\boldsymbol \theta\) if the model expected value at
\(\boldsymbol \theta\) is not very different from the corresponding
model expectation under independence. Using this, it is possible to
investigate what parametric conditions lead to uninterpretability in a
model versus those that guarantee interpretable models. If
\(\text{E}\left[\boldsymbol X | \boldsymbol \theta\right] - \text{E}\left[\boldsymbol X | \boldsymbol \theta^*\right]\)
is large, then the RBM model with parameter vector
\(\boldsymbol \theta\) is said to be uninterpetable. The quantities to
compare in the RBM case are \[
\text{E}\left[\boldsymbol X | \boldsymbol \theta\right] = \sum\limits_{\boldsymbol x \in \mathcal{C}^{\nv + \nh}} \boldsymbol x f_{\boldsymbol \theta}(\boldsymbol x) = \sum\limits_{\boldsymbol x \in \mathcal{C}^{\nv + \nh}} \boldsymbol x \frac{\exp\left(\sum\limits_{i = 1}^\nv \sum\limits_{j=1}^\nh \theta_{ij} v_i h_j + \sum\limits_{i = 1}^\nv\theta_{v_i} v_i + \sum\limits_{j = 1}^\nh\theta_{h_j} h_j\right)}{\sum\limits_{\boldsymbol x \in \mathcal{C}^{\nv + \nh}}\exp\left(\sum\limits_{i = 1}^\nv \sum\limits_{j=1}^\nh \theta_{ij} v_i h_j + \sum\limits_{i = 1}^\nv\theta_{v_i} v_i + \sum\limits_{j = 1}^\nh\theta_{h_j} h_j\right)}
\] and \begin{align*}
\text{E}\left[\boldsymbol X | \boldsymbol 
\theta^*\right] &= \sum\limits_{\boldsymbol x \in \mathcal{C}^{\nv + \nh}} \boldsymbol x \frac{\exp\left(\sum\limits_{i = 1}^\nv \theta_{v_i} v_i + \sum\limits_{j = 1}^\nh\theta_{h_j} h_j\right)}{\sum\limits_{\boldsymbol x \in \mathcal{C}^{\nv + \nh}}\exp\left(\sum\limits_{i = 1}^\nv\theta_{v_i} v_i + \sum\limits_{j = 1}^\nh\theta_{h_j} h_j\right)} \\
\end{align*}

\hypertarget{explorations-of-model-properties-through-simulation}{%
\section{Explorations of model properties through
simulation}\label{explorations-of-model-properties-through-simulation}}

We next explore and numerically explain the relationship between values
of \(\boldsymbol \theta\) and the three notions of model impropriety
(near-degeneracy, instability, and uninterpretability), for RBM models
of varying sizes.

\hypertarget{tiny-example}{%
\subsection{Tiny example}\label{tiny-example}}

\begin{figure}[ht]
  \centering
  \resizebox{1cm}{!}{\input{toymodel.tikz}}
  \caption{A small example restricted Boltzmann machine (RBM), with two nodes, one hidden and one visible.}
  \label{fig:toymodel}
\end{figure}
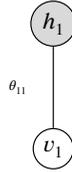

To illustrate the ideas of model near-degeneracy, instability, and
uninterpretability in a RBM, we consider first the smallest possible
(toy) example that consists of one visible node \(v_1\) and one hidden
node \(h_1\) that are both binary. A schematic of this model can be
found in Figure \ref{fig:toymodel}. Because it seems most common, we
shall begin by employing \(0/1\) encoding of binary variables (both
\(h_1\) and \(v_1\) taking values in \(\mathcal{C} = \{0,1\}\)).
(Eventually we shall argue in Section \ref{data-encoding} that \(-1/1\)
coding has advantages.)

\hypertarget{impropriety-three-ways}{%
\subsubsection{Impropriety three ways}\label{impropriety-three-ways}}

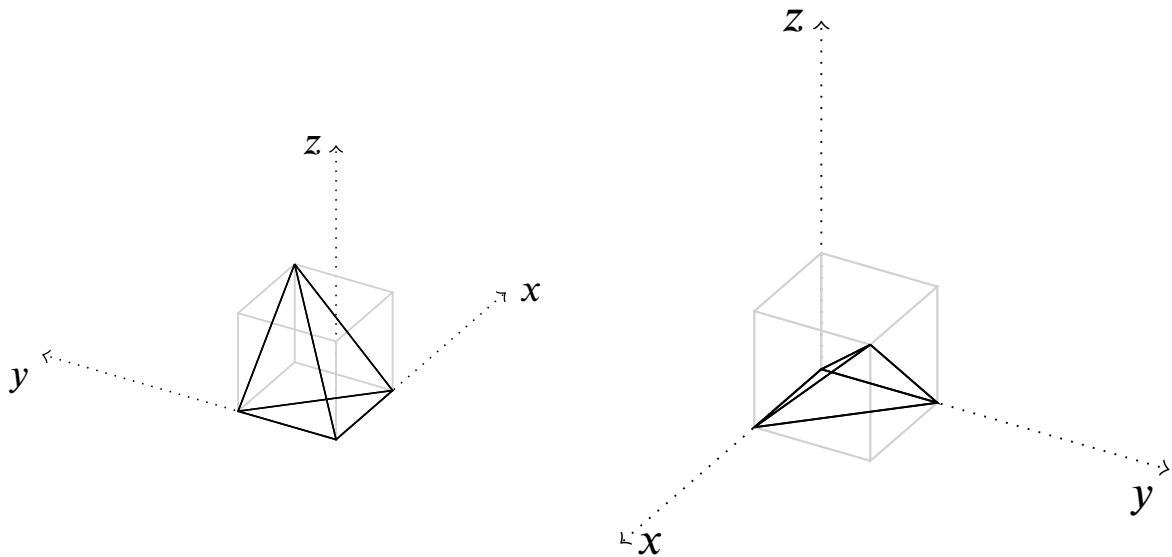
\begin{figure}[ht]
  \begin{minipage}{0.45\textwidth}
  \resizebox{\linewidth}{!}{
    \tdplotsetmaincoords{60}{-60}
    \input{toyhull_top.tikz}
    
    \draw(0,1,0) -- (1,0,0) -- (0,0,0) -- (0,1,0); 
     \draw(1,1,1) -- (1,0,0) -- (0,0,0) -- (1,1,1); 
     \draw(1,1,1) -- (0,1,0) -- (0,0,0) -- (1,1,1); 
     \draw(1,1,1) -- (0,1,0) -- (1,0,0) -- (1,1,1); 
    \end{tikzpicture}
  }
  \end{minipage}
  \begin{minipage}{0.45\textwidth}
  \resizebox{\linewidth}{!}{
    \tdplotsetmaincoords{60}{120}
    \input{toyhull_top.tikz}
    
    \draw(0,1,0) -- (1,0,0) -- (0,0,0) -- (0,1,0); 
     \draw(1,1,1) -- (1,0,0) -- (0,0,0) -- (1,1,1); 
     \draw(1,1,1) -- (0,1,0) -- (0,0,0) -- (1,1,1); 
     \draw(1,1,1) -- (0,1,0) -- (1,0,0) -- (1,1,1); 
    \end{tikzpicture}
  }
  \end{minipage}
  \caption{Two perspectives of the convex hull of the ``statistic space" in three dimensions for the toy RBM with one visible and one hidden node.}
\label{fig:toyhull}
\end{figure}

For this small model, we are able to investigate the symptoms of model
impropriety, beginning with near-degeneracy. To this end, recall from
Section \ref{near-degeneracy} that one characterization requires
consideration of the convex hull of possible values of statistics
\(\boldsymbol t(\boldsymbol x)\), \[
\mathcal{T} = \{\boldsymbol t(\boldsymbol x): \boldsymbol x = (v_1, h_1) \in \{0,1\}^2\} \equiv \{(v_1, h_1, v_1 h_1): v_1, h_1 \in \{0,1\}\}
\] appearing in the RBM probabilities for this model. As this set is in
three dimensions, we are able to explicitly illustrate the shape of
boundary of the convex hull of \(\mathcal{T}\) and explore the behavior
of the mean vector
\(\boldsymbol \mu(\boldsymbol \theta) = \text{E}_{\boldsymbol \theta} \boldsymbol t(\boldsymbol x)\)
as a function of the parameter vector \(\boldsymbol \theta\). Figure
\ref{fig:toyhull} shows the convex hull of our ``statistic space,''
\(\mathcal{T} \subset \{0,1\}^3\), for this toy problem from two
perspectives (enclosed by the unit cube \([0,1]^3\), the convex hull of
\(\{0,1\}^3\)). In this small model, note that the convex hull of
\(\mathcal{T}\) does not fill the unrestricted hull of \(\{0,1\}^3\)
because of the relationship between the elements of
\(\mathcal{T} = \{(v_1, h_1, v_1 h_1 : v_1, h_1 \in \{0, 1\} \}\) (i.e.
\(v_1 h_1 = 1\) only if \(v_1 = h_1 = 1\)).

\par

We can compute the mean vector for \(\boldsymbol t(\boldsymbol x)\) as a
function of the model parameters as \begin{align*}
\boldsymbol \mu(\boldsymbol \theta) = \text{E}_{\boldsymbol \theta}\left[ \boldsymbol t(\boldsymbol X) \right] =  \sum\limits_{\boldsymbol x = (v_1, h_1) \in \{0,1\}^2} \left\{ t(x) \frac{\exp\left( \theta_{11} h_1 v_1 + \theta_{h1} h_1 + \theta_{v1} v_1\right)}{\gamma(\boldsymbol \theta)} \right\} = \left[
\begin{matrix}
\frac{\exp\left(\theta_{v_1}\right) + \exp\left(\theta_{11} + \theta_{v_1} + \theta_{h_1}\right)}{\gamma(\boldsymbol \theta)} \\
\frac{\exp\left(\theta_{h_1}\right) + \exp\left(\theta_{11} + \theta_{v_1} + \theta_{h_1}\right)}{\gamma(\boldsymbol \theta)} \\
\frac{\exp\left(\theta_{11} + \theta_{v_1} + \theta_{h_1}\right)}{\gamma(\boldsymbol \theta)} \\
\end{matrix} \right]
\end{align*} where
\(\gamma(\boldsymbol \theta) = \sum\limits_{h_1 = 0}^1\sum\limits_{v_1 = 0}^1 \exp(\theta_{11} h_1 v_1 + \theta_{h_1}h_1 + \theta_{v_1}v_1)\).
The three parametric coordinate functions of
\(\boldsymbol \mu(\boldsymbol \theta)\) can be represented as in Figure
\ref{fig:degen-toy}. (Contour plots for three coordinate functions are
shown in columns for various values of \(\theta_{11}\), which can be
interpreted here as an absolute log-odds ratio as the visible changes
between \(0\) and \(1\).) In examining these, we see that as coordinates
of \(\boldsymbol \theta\) grow larger in magnitude, at least one mean
function for the entries of \(\boldsymbol t(\boldsymbol x)\) approaches
a value 0 or 1, forcing
\(\boldsymbol \mu(\boldsymbol \theta) = \text{E}_{\boldsymbol \theta} \boldsymbol t(\boldsymbol x)\)
to be near to the boundary of the convex hull of \(\mathcal{T}\), as a
sign of model near-degeneracy. Thus, for a very small example we can see
the relationship between values of \(\boldsymbol \theta\) moving
(sometimes only slightly) from zero and model degeneracy.

\par

\begin{figure}

{\centering \includegraphics{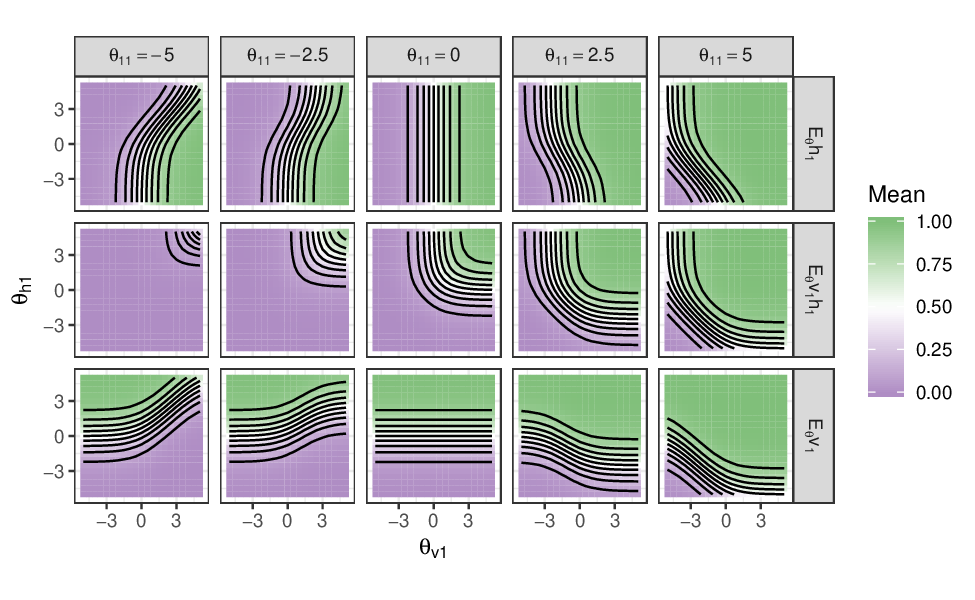} 

}

\caption{\label{fig:degen-toy}Contour plots for the three parametric mean functions of sufficient statistics for a RBM with one visible and one hidden node.}\label{fig:degen-toy}
\end{figure}

Secondly, we can look at \(\text{LREP}(\boldsymbol \theta)\) from
\eqref{eq:elpr} for this tiny model in order to consider model
instability as a function of RBM parameters. Recall that large values of
\(\text{LREP}(\boldsymbol \theta)\) are associated with an extreme
sensitivity of the model probabilities
\(f_{\boldsymbol \theta}(\boldsymbol x)\) to small changes in
\(\boldsymbol x\) (see Proposition \ref{prp:instab1}). The quantity
\(\text{LREP}(\boldsymbol \theta)\) for this small RBM is \begin{align*}
\text{LREP}(\boldsymbol \theta) &= \log \left[\frac{\max\limits_{(v_1, \dots, v_\nv) \in \mathcal{C}^\nv}P_{\boldsymbol \theta_\nv}(v_1, \dots, v_\nv)}{\min\limits_{(v_1, \dots, v_\nv) \in \mathcal{C}^\nv}P_{\boldsymbol \theta_\nv}(v_1, \dots, v_\nv)}\right] = \log \left[\frac{\max\limits_{v_1 \in \mathcal{C}}\sum\limits_{h_1 \in \mathcal{C}}\exp\left\{\theta_{11} h_1 v_1 + \theta_{h_1}h_1 + \theta_{v_1} v_1 \right\}}{\min\limits_{v_1 \in \mathcal{C}}\sum\limits_{h_1 \in \mathcal{C}}\exp\left\{\theta_{11} h_1 v_1 + \theta_{h_1}h_1 + \theta_{v_1} v_1 \right\}}\right]. \\
\end{align*}

Figure \ref{fig:instab} shows contour plots of
\(\text{LREP}(\boldsymbol \theta)/\nv\) for various values of
\(\boldsymbol \theta\) in this model with \(\nv = 1\). We can see that
this quantity is large for large magnitudes of \(\boldsymbol \theta\),
especially for large values of the dependence/interaction parameter
\(\theta_{11}\). This suggests instability as \(|\boldsymbol \theta|\)
becomes large, agreeing also with the concerns about near-degeneracy
produced by consideration of \(\boldsymbol \mu(\boldsymbol \theta)\).

\par

\begin{figure}

{\centering \includegraphics{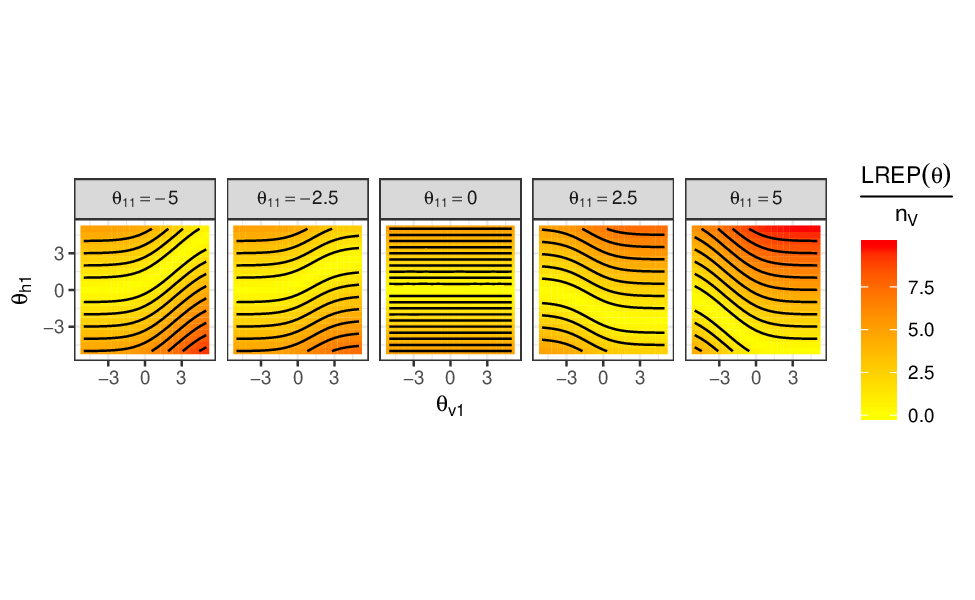} 

}

\caption{$\label{fig:instab}\text{LREP}(\boldsymbol \theta)/\nv$ for various values of $\boldsymbol \theta$ for the tiny example model. (Recall here $\nv$ is the number of visible nodes and here is $1$.) This quantity is large for large magnitudes of $\boldsymbol \theta$.}\label{fig:instab}
\end{figure}

Finally to consider the effect of \(\boldsymbol \theta\) on potential
model uninterpretability, we can look at the difference between model
expectations, E\(\left[\boldsymbol X | \boldsymbol \theta\right]\), and
expectations given independence,
E\(\left[\boldsymbol X | \boldsymbol \theta^*\right ]\) for the tiny toy
RBM model where \(\boldsymbol X = (V_1, H_1, V_1 H_1)\). This difference
is given by \begin{align*}
\text{E}\left[\boldsymbol X | \boldsymbol \theta\right] -   \text{E}\left[\boldsymbol X | \boldsymbol \theta^* \right] 
&= \left[
\begin{matrix}
\frac{\exp\left(\theta_{11} + \theta_{v_1} + 2\theta_{h_1}\right) - \exp\left( \theta_{v_1} + 2\theta_{h_1}\right) }{\left(\exp\left(\theta_{v_1}\right) + \exp\left(\theta_{h_1}\right) + \exp\left(\theta_{11} + \theta_{v_1} + \theta_{h_1}\right)\right)\left(\exp\left(\theta_{v_1}\right) + \exp\left(\theta_{h_1}\right) + \exp\left(+ \theta_{v_1} + \theta_{h_1}\right)\right)} \\
\frac{\exp\left(\theta_{11} + 2\theta_{v_1} + \theta_{h_1}\right) - \exp\left( 2\theta_{v_1} + \theta_{h_1}\right)  }{\left(\exp\left(\theta_{v_1}\right) + \exp\left(\theta_{h_1}\right) + \exp\left(\theta_{11} + \theta_{v_1} + \theta_{h_1}\right)\right)\left(\exp\left(\theta_{v_1}\right) + \exp\left(\theta_{h_1}\right) + \exp\left(+ \theta_{v_1} + \theta_{h_1}\right)\right)}
\end{matrix}\right].
\end{align*}

Again, we can inspect these coordinate functions of this vector
difference to look for a relationship between parameter values and large
values of
\(\text{E}[\boldsymbol X|\boldsymbol \theta] - \text{E}[\boldsymbol X| \boldsymbol \theta^*]\)
as a signal of uninterpretability for the toy RBM.

\par

\begin{figure}

{\centering \includegraphics{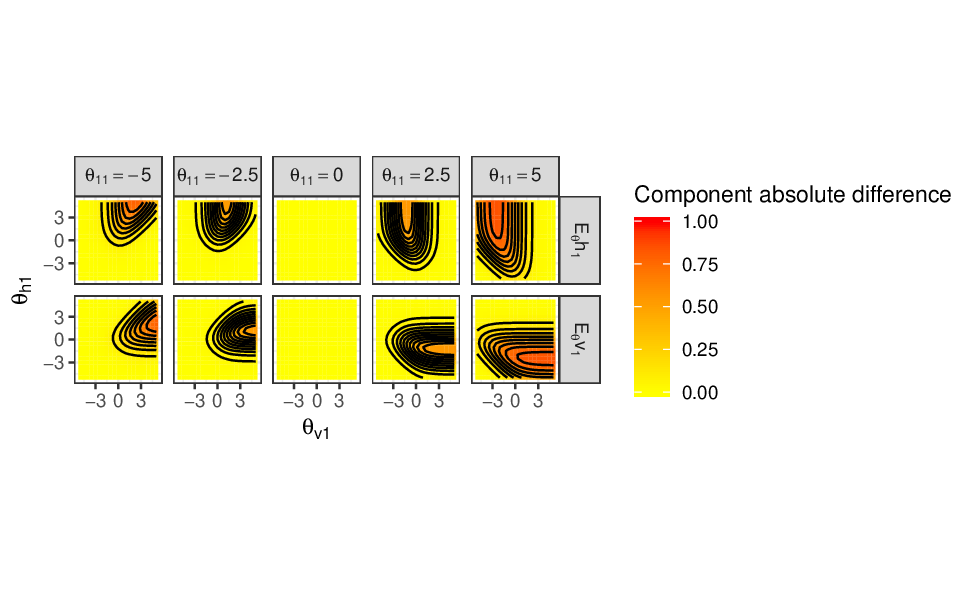} 

}

\caption{\label{fig:uninterp}The absolute difference between coordinates of model expectations, E$\left[\boldsymbol X | \boldsymbol \theta\right]$, and expectations given independence, E$\left[\boldsymbol X | \boldsymbol \theta^* \right ]$ for a RBM with one visible and one hidden node. As an indicator of uninterpretability, note that differences in expectations increase as the dependence parameter $\theta_{11}$ deviates from zero.}\label{fig:uninterp}
\end{figure}

Figure \ref{fig:uninterp} shows that the absolute difference between
coordinates of the vector of model expectations,
E\(\left[\boldsymbol X | \boldsymbol \theta\right]\) and corresponding
expectations E\(\left[\boldsymbol X | \boldsymbol \theta^*\right ]\)
given independence grow for the toy RBM as the values of
\(\boldsymbol \theta\) are farther from zero, especially for large
magnitudes of the dependence parameter \(\theta_{11}\). This is a third
indication that parameter vectors of large magnitude can readily lead to
model impropriety in a RBM.

\hypertarget{data-encoding}{%
\subsubsection{Data encoding}\label{data-encoding}}

Multiple encodings of the binary variables are possible. For example, we
could allow hiddens \((H_1, \dots, H_\nh) \in \{0,1\}^\nh\) and visibles
\((V_1, \dots, V_\nv) \in \{0,1\}^\nv\), as in the previous sections or
we could instead encode the state of the variables as \(\{-1,1\}^\nh\)
and \(\{-1,1\}^\nv\). This will result in variables
\(\boldsymbol t(\boldsymbol X)\) from \eqref{eq:t} satisfying
\(\boldsymbol t(\boldsymbol x) \in \{0,1\}^{\nh + \nv + \nh*\nv}\) or
\(\boldsymbol t(\boldsymbol x) \in \{-1,1\}^{\nh + \nv + \nh*\nv}\)
depending on how we encode ``on'' and ``off'' states in the nodes.

The \(-1/1\) data encoding has the benefit of providing a
guaranteed-to-be non-degenerate model at
\(\boldsymbol \theta = \boldsymbol 0 \in \mathbb{R}^{\nh + \nv + \nh*\nv}\),
where the zero vector then serves as the natural center of the parameter
space and induces the simplest possible model properties for the RBM
(i.e., at \(\boldsymbol \theta = \boldsymbol 0\), all variables are
independent uniformly distributed on \(\{-1,1\}^\nv\)). The proof of
this and further exploration of the equivalence of the
\(\boldsymbol \theta\) parameterization of the RBM model class and
parameterization by \(\boldsymbol \mu(\boldsymbol \theta)\) is in the
on-line supplementary materials. Hence, while from some computing
perspectives \(0/1\) coding might seem most natural, the \(-1/1\) coding
is far more convenient and interpretable from the point of view of
statistical modeling, where it makes parameters simply interpreted in
terms of symmetrically defined main effects and interactions. Under the
data encoding \(-1/1\), the parameter space centered at 0 is also
helpful for framing parameter configurations that are undesirably large
(i.e., these are naturally parameters that have moved too far away from
\(\bm{0}\) where the RBM model is anchored to be trivially describable
and completely problem-free). In light of all of these matters we will
henceforth employ the \(-1/1\) coding.

\hypertarget{exploring-manageable-examples}{%
\subsection{Exploring manageable
examples}\label{exploring-manageable-examples}}

To explore the impact of RBM parameter vector \(\boldsymbol \theta\)
magnitude on near-degeneracy, instability, and uninterpretability, we
consider models of small size. For \(\nh, \nv \in \{1, \dots, 4\}\), we
sample 100 values of \(\boldsymbol \theta\) with various magnitudes
(details to follow). For each set of parameters we then calculate
metrics of model impropriety introduced in Section
\ref{degeneracy-instability-and-uninterpretability} based on
\(\boldsymbol \mu(\boldsymbol \theta)\),
\(\text{LREP}(\boldsymbol \theta)/\nv\), and the absolute coordinates of
\(\text{E}\left[\boldsymbol X | \boldsymbol \theta\right] - \text{E}\left[\boldsymbol X | \boldsymbol \theta^* \right]\).
In the case of near-degeneracy, we classify each model as
near-degenerate based on the distance of
\(\boldsymbol \mu(\boldsymbol \theta)\) from the boundary of the convex
hull of \(\mathcal{T}\) and look at the fraction of models that are
``near-degenerate,'' meaning they are within a small distance
\(\epsilon > 0\) of the boundary of the convex hull. We define ``small''
through a rough estimation of the volume of the hull for each model
size. We pick \(\epsilon_0 = 0.05\) for \(\nh=\nv=1\) and then, for
every other \(\nh\) and \(\nv\), set \(m=\nh+\nv+\nv*\nh\) and pick
\(\epsilon\) so that \(1-(1-2\epsilon_0)^3 = 1 - (1-2\epsilon)^m\). In
this way, we roughly scale the volume of the ``small distance'' to the
boundary of the convex hull to be equivalent across model dimensions.

In our numerical experiment, we split
\(\boldsymbol \theta = (\boldsymbol \theta_{main}, \boldsymbol \theta_{interaction})\)
into \(\boldsymbol \theta_{main}\) and
\(\boldsymbol \theta_{interaction}\), in reference to which variables in
the probability function the parameters correspond (whether they
multiply a \(v_i\) or a \(h_j\) or they multiply a \(v_i h_j\)), and
allow the two types of terms to have varying average magnitudes,
\(||\boldsymbol \theta_{main} || /(\nh+\nv)\) and
\(||\boldsymbol \theta_{interaction} || /(\nh*\nv)\). These average
magnitudes vary on a \(1\)-dimensional grid between 0.001 and 3 with 24
breaks, yielding 576 \(2\)-dimensional grid points. (By looking at the
average magnitudes, we are able to later consider the potential benefit
of shrinking each parameter value \(\theta_i\) towards zero in a
Bayesian fitting technique.) At each point in the grid, 100 vectors
(\(\boldsymbol \theta_{main}\)) are sampled uniformly on a sphere with
radius corresponding to the first coordinate of the point and 100
vectors (\(\boldsymbol \theta_{interction}\)) are sampled uniformly on a
sphere with radius corresponding to the second coordinate of the point
via sums of squared and scaled iid Normal\((0, 1)\) variables. These
vectors are then paired to create 100 values of \(\boldsymbol \theta\)
at each point in the grid.

\begin{figure}
\centering
\includegraphics{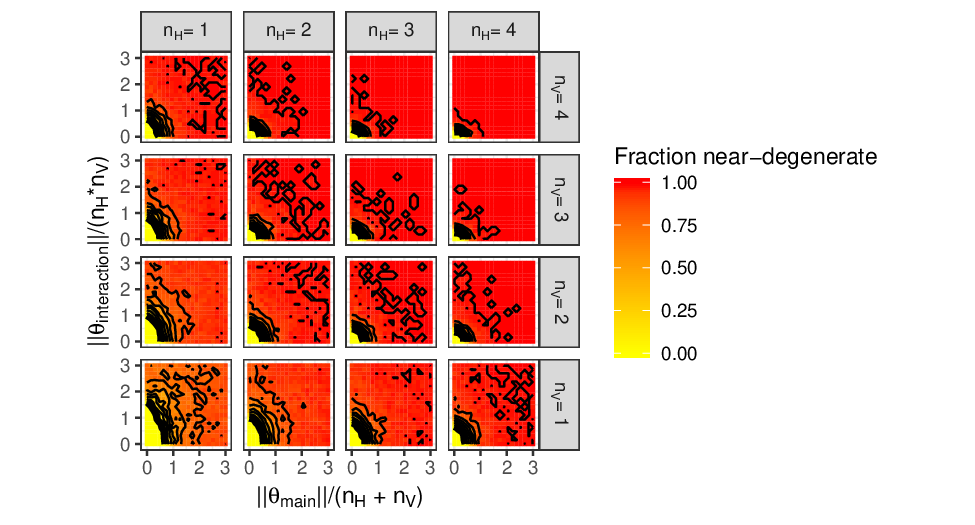}
\caption{\label{fig:degen-plots}Results from the numerical experiment,
here looking at the fraction of models that were near-degenerate for
each combination of magnitudes of the points \(\boldsymbol \theta\) and
model size, where
\(\boldsymbol \theta = (\boldsymbol \theta_{main}, \theta_{interaction})\)
is split into main and interaction parameters. Black lines show the
contour levels for fraction of near-degeneracy, while the thick black
line shows the level where the fraction of near-degenerate models is
.05.}
\end{figure}

\begin{figure}
\centering
\includegraphics{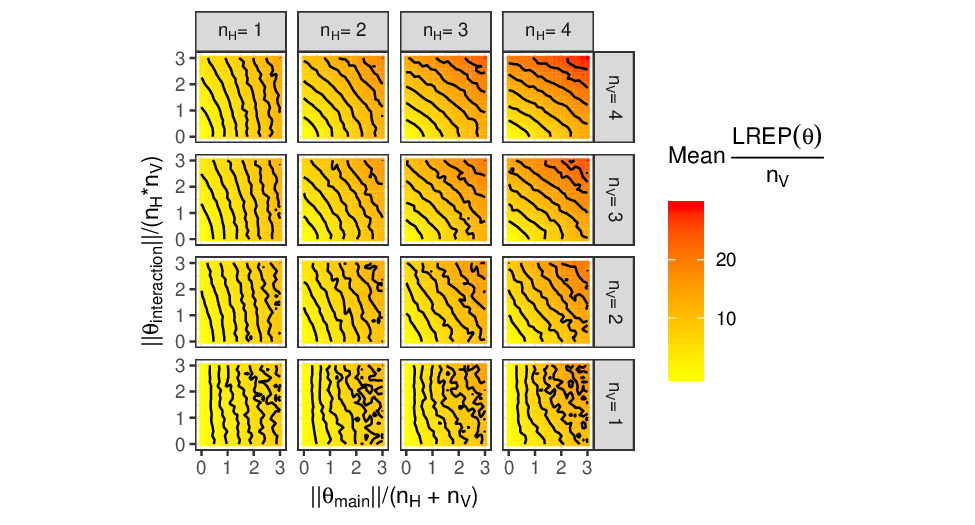}
\caption{\label{fig:instab-plots}The sample mean value of
\(\text{LREP}(\boldsymbol \theta)/\nv\) at each grid point for each
combination of magnitudes of the points of \(\boldsymbol \theta\) and
model size. As the magnitude of \(\boldsymbol \theta\) grows, so does
the value of this metric, indicating typical instability in the model.}
\end{figure}

\begin{figure}
\centering
\includegraphics{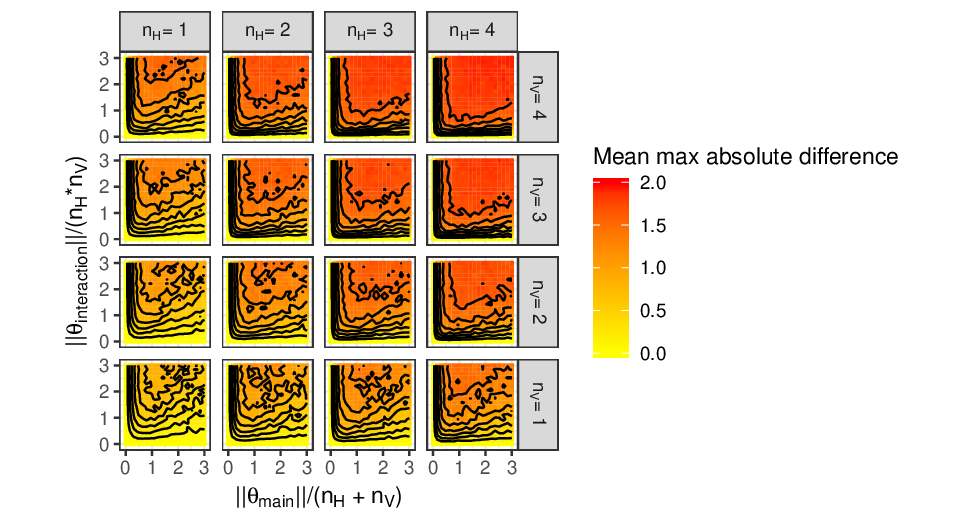}
\caption{\label{fig:uninterp-plots}The sample mean of the maximum
component of the absolute difference between the model expectation
vector, E\(\left[\boldsymbol X | \boldsymbol \theta\right]\), and the
expectation vector given independence,
E\(\left[\boldsymbol X | \boldsymbol \theta^* \right ]\). Larger
magnitudes of \(\boldsymbol \theta\) correspond to larger differences,
thus indicating reduced interpretability.}
\end{figure}

The results of this numerical study are summarized in Figures
\ref{fig:degen-plots}, \ref{fig:instab-plots}, and
\ref{fig:uninterp-plots}. From these three figures, it is clear that all
three measures of model impropriety show higher values for larger
magnitudes of the parameter vectors, supporting the RBM model properties
developed in Section \ref{background}. As a compounding issue, these
figures show that, as models grow in size, it becomes easier for more
parameter configurations to push RBM models into near-degeneracy,
instability and uninterpretability. Additionally, since there are
\(\nh*\nv\) interaction terms in \(\boldsymbol \theta\) versus only
\(\nh + \nv\) main effect terms, for large models there are many more
interaction parameters than main effects in the models. And so, severely
limiting the magnitude of the individual interactions may well help
prevent model impropriety.

\par

\begin{figure}
\centering
\includegraphics{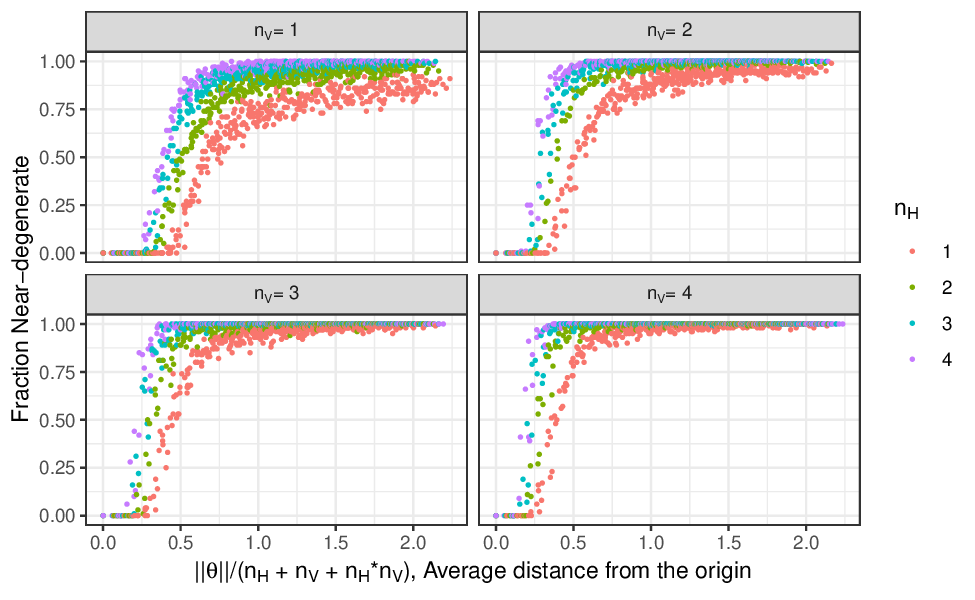}
\caption{\label{fig:split-plots-dist}The fraction of near-degenerate
models for each magnitude of \(\boldsymbol \theta\). For each number
\(\nv\) of visibles in the model, the fraction near-degenerate moves
away from zero at larger values of \(||\boldsymbol \theta||\) as the
number \(\nh\) of hidden variables increases and the slope becomes
steeper as \(\nh\) increases as well.}
\end{figure}

Figure \ref{fig:split-plots-dist} shows the fraction of near-degenerate
models for each magnitude of \(\boldsymbol \theta\) for each model
architecture. For each number \(\nv\) of visibles in the model, as the
number \(\nh\) of hiddens increase, the fraction near-degenerate
diverges from zero at increasing rates for larger values of
\(||\boldsymbol \theta||\). This shows that, as model size gets larger,
the risk of degeneracy starts at a slightly larger magnitude of
parameters, but very quickly increases until reaching close to 1.

\par

These manageable examples indicate that RBMs are near-degenerate,
unstable, and uninterpretable for large portions of the parameter space
having large \(\|\boldsymbol \theta\|\). These problematic aspects
require serious consideration when using RBM models, on top of the
additional matter of principled/rigorous fitting of RBM models.

\hypertarget{model-fitting}{%
\section{Model Fitting}\label{model-fitting}}

Typically, fitting a RBM via maximum likelihood (ML) methods will be
unfeasible due mainly to the intractability of the normalizing term
\(\gamma(\boldsymbol \theta)\) in a model \eqref{eq:pmf} of any
realistic size. Ad hoc methods are often recommended instead, which aim
to avoid this problem by using stochastic ML approximations that employ
a small number of MCMC draws (i.e., contrastive divergence, (Hinton
2002)).

However, computational concerns are not the only issues with fitting a
RBM using ML. In addition, a RBM model, with the appropriate choice of
parameters and number of hiddens, has the potential to re-create any
distribution for the data (i.e., reproduce any specification of cell
probabilities for the binary data outcomes). For example, Montufar and
Ay (2011) show that any distribution on \(\{0, 1\}^{\nv}\) can be
approximated arbitrarily well by a RBM with \(2^{\nv-1} - 1\) hidden
units. We provide a small example in the on-line supplementary materials
that illustrates such approximations.

Furthermore, as development in the online appendix shows, not only is it
possible to approximate any distribution on the visibles arbitrarily
well (cf. Montufar and Ay 2011), but dramatically different parameter
settings can induce the same RBM model (beyond mere symmetries in
parameterization). A further (and somewhat odd) consequence of the RBM
parameterization is then that when fitting the RBM model by
likelihood-based methods, we may already know the nature of the answer
before we begin: namely, such fitting will simply reproduce the
empirical distribution of the training data if sufficiently many hiddens
are in the model. There then can be no model refinements or smoothing
from the RBM. That is, based on a random sample of vectors of visible
variables, the model for the cell probabilities with the highest
likelihood over \emph{all possible model classes} (i.e., RBM-based or
not) is the empirical distribution, and the over-parameterization of the
RBM model itself ensures that this empirical distribution can be
arbitrarily well-approximated. Not only does RBM model fitting based on
ML seek to reproduce the empirical distribution, whenever this empirical
distribution contains empty cells, fitting steps for the RBM model will
further aim to choose parameters that necessarily diverge to infinity in
magnitude in order to zero-out the corresponding RBM cell probabilities.
In data applications with a large sample space, it is unlikely that the
training set will include at least one of each possible vector outcome
(unlike this small example). This implies that some RBM model parameters
must diverge to \(+\infty\) to mimic the empirical distribution with
empty cells and, as we have already discussed in Section
\ref{explorations-of-model-properties-through-simulation}, large
magnitudes of \(\boldsymbol \theta\) lead to model impropriety in the
RBM.

Here we consider what might be done in a principled manner to prevent
both overfitting and model impropriety, testing on a \(\nv = \nh = 4\)
case that already stretches the limits of what is computable --- in
particular we consider Bayes methods. We employ a Bayesian framework
because of the ability to obtain uncertainty quantification with little
to no extra effort, via posterior credible intervals.

\hypertarget{bayesian-model-fitting}{%
\subsection{Bayesian model fitting}\label{bayesian-model-fitting}}

To avoid model impropriety for a fitted RBM, we wish to avoid parts of
the parameter space \(\mathbb{R}^{\nv + \nh + \nv*\nh}\) that lead to
near-degeneracy, instability, and uninterpretability. Motivated by the
insights in Section \ref{exploring-manageable-examples}, one idea is to
shrink
\(\boldsymbol \theta = (\boldsymbol \theta_{main}, \boldsymbol \theta_{interaction})\)
toward \(\boldsymbol 0\) by specifying priors that place low probability
on large values of \(||\boldsymbol \theta||\), specifically focusing on
shrinking \(\boldsymbol \theta_{interaction}\) more than
\(\boldsymbol \theta_{main}\). This is similar to an idea advocated by
Hinton (2010) called \emph{weight decay}, in which a penalty is added to
the interaction terms in the model,
\(\boldsymbol \theta_{interaction}\), shrinking their magnitudes.

\par

\begin{table}[ht]
\centering
\begin{tabular}{rrrrrr}
  \hline
Parameter & Value & Parameter & Value & Parameter & Value \\ 
  \hline
$\theta_{v1}$ & $-1.104376$ & $\theta_{11}$ & $-0.0006334$ & $\theta_{31}$ & $-0.0038301$ \\ 
  $\theta_{v2}$ & $-0.2630044$ & $\theta_{12}$ & $-0.0021401$ & $\theta_{32}$ & $0.0032237$ \\ 
  $\theta_{v3}$ & $0.3411915$ & $\theta_{13}$ & $0.0047799$ & $\theta_{33}$ & $0.0020681$ \\ 
  $\theta_{v4}$ & $-0.2583769$ & $\theta_{14}$ & $0.0025282$ & $\theta_{34}$ & $0.0041429$ \\ 
  $\theta_{h1}$ & $-0.1939302$ & $\theta_{21}$ & $0.0012975$ & $\theta_{41}$ & $0.0089533$ \\ 
  $\theta_{h2}$ & $-0.0572858$ & $\theta_{22}$ & $0.0000253$ & $\theta_{42}$ & $-0.0042403$ \\ 
  $\theta_{h3}$ & $-0.2101802$ & $\theta_{23}$ & $-0.0004352$ & $\theta_{43}$ & $-0.000048$ \\ 
  $\theta_{h4}$ & $0.2402456$ & $\theta_{24}$ & $-0.0086621$ & $\theta_{44}$ & $0.0004767$ \\ 
   \hline
\end{tabular}
\caption{Parameters used to fit a test case with $V = H = 4$. This parameter vector was chosen as a sampled value of $\boldsymbol \theta$ that was not near the convex hull of the sufficient statistics for a grid point in Figure \ref{fig:degen-plots} with $< 5$\% near-degeneracy.} 
\label{tab:theta}
\end{table}

We considered a test case with \(\nv = \nh = 4\) and parameters given in
Table \ref{tab:theta}. This parameter vector was chosen as a sampled
value of \(\boldsymbol \theta\) at which the resulting RBM model would
not be clearly degenerate. We simulated \(n = 5,000\) realizations of
visibles as a training set and fit the RBM using three Bayes
methodologies. These involved the following set-ups for choice of prior
distribution \(\pi(\boldsymbol \theta)\) for parameters
\(\boldsymbol \theta\).

\begin{enumerate}
\def\labelenumi{\arabic{enumi}.}
\tightlist
\item
  \emph{A ``trick'' prior.} Here we cancel out normalizing term in the
  likelihood (from \(\gamma(\boldsymbol \theta)\) in \eqref{eq:pmf}) so
  that resulting full conditionals of \(\boldsymbol \theta\) are
  multivariate Normal. Namely, this involves a prior of the form
  \begin{align*}
   \pi(\boldsymbol \theta) \propto \gamma(\boldsymbol \theta)^n \exp\left(-\frac{1}{2C_{1}}\boldsymbol \theta_{main}'\boldsymbol \theta_{main} -\frac{1}{2C_{2}}\boldsymbol \theta_{interaction}'\boldsymbol \theta_{interaction}\right), \vspace{-.75cm}
   \end{align*} where \begin{align*}
   \gamma(\boldsymbol \theta) = \sum\limits_{\boldsymbol x \in \mathcal{C}^{\nh + \nv}}\exp\left(\sum\limits_{i = 1}^\nv\sum\limits_{j=1}^\nh \theta_{ij} v_i h_j + \sum\limits_{i = 1}^V\theta_{v_i} v_i + \sum\limits_{j = 1}^\nh\theta_{h_j} h_j\right) 
   \end{align*} for hyperparameter choices \(0< C_2 < C_1\). The unknown
  hidden variables \(h_j\) are also directly treated as latent variables
  and are sampled in each MCMC iterative draw from the posterior
  distribution. This is the method of Li (2014). We will refer to this
  method as Bayes with Trick Prior and Latent Variables (BwTPLV).
\item
  \emph{A truncated Normal prior.} Here we use independent spherical
  normal distributions as priors for \(\boldsymbol \theta_{main}\) and
  \(\boldsymbol \theta_{interaction}\), which are \emph{truncated} at
  \(3\sigma_{main}\) and \(3\sigma_{interaction}\), respectively, based
  on standard deviation hyperparameters
  \(0<\sigma_{interaction}<\sigma_{main}\). Full conditional
  distributions are not conjugate, and simulation from the posterior was
  accomplished using a geometric adaptive Metropolis Hastings step (Zhou
  2014) and calculation of likelihood normalizing constant. (This
  computation is barely feasible for a problem of this size and would be
  unfeasible for larger problems.) Here the hidden variables \(h_j\) are
  again carried along in the MCMC implementation as latent variables. We
  will refer to this method as Bayes with Truncated Normal prior and
  Latent Variables (BwTNLV).
\item
  \emph{A truncated Normal prior and marginalized likelihood.} Here we
  marginalize out the hidden variables
  \(\boldsymbol h = (h_1, \dots, h_\nh)\) in
  \(f_{\boldsymbol \theta}(\boldsymbol x)\), and use the truncated
  Normal priors applied to the marginal probabilities for visible
  variables given by \begin{align*}
   g_{\boldsymbol \theta}(\boldsymbol v) \propto \sum\limits_{\boldsymbol h \in \mathcal{C}^\nh} \exp\left(\sum\limits_{i = 1}^\nv \sum\limits_{j=1}^\nh \theta_{ij} v_i h_j + \sum\limits_{i = 1}^\nv\theta_{v_i} v_i + \sum\limits_{j = 1}^\nh\theta_{h_j} h_j\right), \boldsymbol v \in \mathcal{C}^\nv.
   \end{align*} We will refer to this method as Bayes with Truncated
  Normal prior and Marginalized Likelihood (BwTNML).
\end{enumerate}

The three fitting methods are ordered above according to computational
feasibility in a real-data situation, with BwTPLV being the most
computationally feasible due to conjugacy and BwTNML the least feasible
due to the marginalization and need for an adaptive Metropolis Hastings
step. All three methods require choosing the values of hyperparameters.
In each case, we have chosen these values based on a rule of thumb that
shrinks \(\boldsymbol \theta_{interaction}\) more than
\(\boldsymbol \theta_{main}\). Additionally, BwTPLV requires additional
tuning (i.e.~a tuning parameter \(C > 0\) in Table \ref{tab:hyperparam})
to choose \(C_1\) and \(C_2\), reducing its appeal. The forms used for
the hyperparameters in our simulation are presented in Table
\ref{tab:hyperparam}.

\begin{table}[ht]
\centering
\begin{tabular}{|l|c|c|}
\hline 
Method & Hyperparameter & Value \\ 
\hline \hline
\multirow{2}{*}{BwTPLV} & $C_1$ & $\frac{C}{n}\frac{1}{\nh + \nv}$ \\
 & $C_2$ & $\frac{C}{n}\frac{1}{\nh*\nv}$ \\
\hline
\multirow{2}{*}{BwTNLV} & $\sigma^2_{main}$ & $\frac{1}{\nh + \nv}$ \\
 & $\sigma^2_{interaction}$ & $\frac{1}{\nh*\nv}$ \\
\hline
\multirow{2}{*}{BwTNML} & $\sigma^2_{main}$ & $\frac{1}{\nh + \nv}$ \\
 & $\sigma^2_{interaction}$ & $\frac{1}{\nh*\nv}$ \\
\hline
\end{tabular}
\caption{The values used for the hyperparameters for all three fitting methods. A rule of thumb is imposed which decreases prior variances for the model parameters as the size of the model increases and also shrinks $\boldsymbol \theta_{interaction}$ more than $\boldsymbol \theta_{main}$. The common $C$ defining $C_1$ and $C_2$  in the BwTPLV method is chosen by tuning.}
\label{tab:hyperparam}
\end{table}

\par

\begin{figure}
\centering
\includegraphics{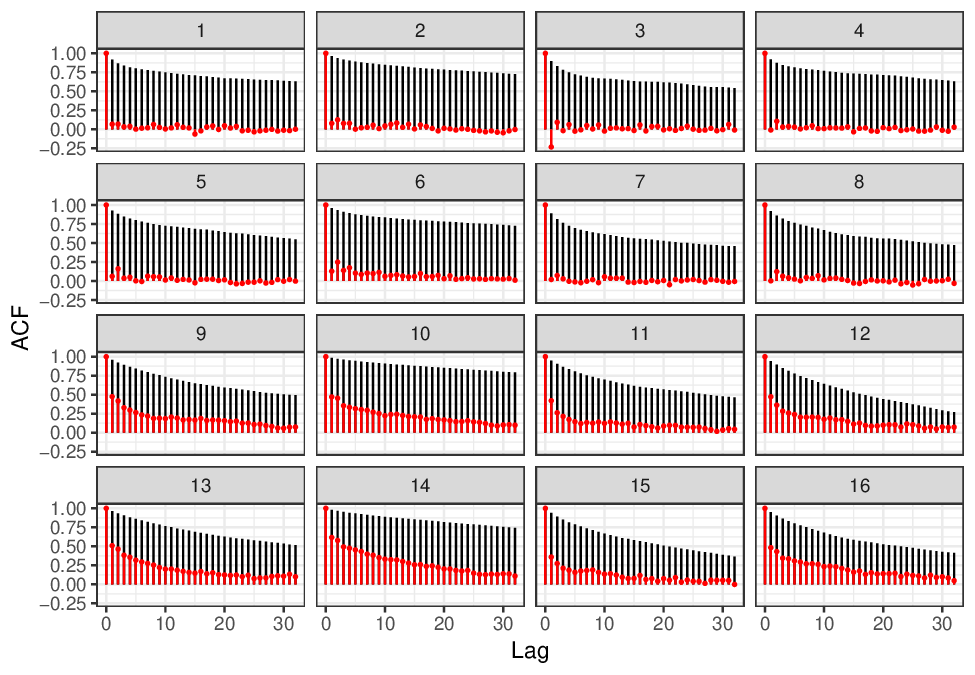}
\caption{\label{fig:acf}The autocorrelation functions (ACF) for the
posterior probabilityies of all \(2^4 = 16\) possible outcomes for the
vector of four visibles assessed at multiple lags for each method with
BwTNLV in black and BwTNML in red. As expected, ACF corresponding to the
method that marginalizes out the hidden variables from the likelihood
decreases to zero at a much faster rate, indicating better mixing for
the chain.}
\end{figure}

It should be noted that, due to the common prior distributions for
\(\boldsymbol \theta\), both BwTNLV (method 2 above) and BwTNML (method
3) are drawing from the same stationary posterior distribution for
vectors of visibles. A fundamental difference between these two methods
lies in how well these two chains mix and how quickly they arrive at the
target posterior distribution. After a burn-in of 50 iterations selected
by inspecting the trace plots, we assess the issue of mixing in two
ways. First, the autocorrelation functions (ACF) from each posterior
sample corresponding to a model probability for a visible vector outcome
\(\mathbf{v}=(v_1,v_2,v_3,v_4)\in\{\pm 1\}^4\) (i.e., computed from
\(\boldsymbol \theta\) under \eqref{eq:pmf} are determined and plotted
in Figure \ref{fig:acf} with BwTNLV in black and BwTNML in red. As
expected, ACF corresponding to the method (BwTNML) that marginalizes out
the hidden variables from the likelihood decreases to zero at a much
faster rate, indicating better mixing for the chain.

\par

Secondly, we can assess the mixing of the BwTNLV/BwTNML chains using the
notion of effective sample size. If the MCMC chain were truly iid draws
from the target distribution, then for the parameter \(p^{(i)}\)
denoting the probability of the \(i\)th vector outcome for the four
visibles \(\mathbf{v}=(v_1,v_2,v_3,v_4)\in\{\pm 1\}^4\),
\(i=1,\ldots,16\), its estimate as the average \(\bar{p}^{(i)}\) of
posterior sample versions would be approximately Normal with mean given
by the posterior marginal mean of \(p^{(i)}\), and variance given by
\(\sigma^2_i/M\), where \(\sigma^2_i\) is the true posterior variance of
\(p^{(i)}\) and \(M\) is the length of the chain. However, with the
presence of correlation in our chain, the asymptotic variance of
\(\bar{p}^{(i)}\) is instead approximately some \(C_i/M\), where \(C_i\)
is some positive constant such that \(C_i > \sigma^2_i\). We can use an
overlapping block-means approach (Gelman, Shirley, and others 2011) to
get a crude estimate for \(C_i\) as \(\hat{C}_i = bS_b^2\), where
\(S_b^2\) denotes the sample variance of overlapping block means
\(\{\bar{p}_j^{(i)}=\sum_{k=j}^{j+b-1} p_k^{(i)}/b\}_{j=1}^{M-b+1}\) of
length \(b\) computed from the posterior samples
\(\{p_k^{(i)}\}_{k=1}^M\). We compare it to an estimate of
\(\sigma^2_i\) using sample variance \(\hat{\sigma}^2_i\) of the raw
chain, \(\{p_k^{(i)}\}_{k=1}^{M}\). Formally, we approximate the
effective sample size of the length \(M\) MCMC chain as \[
M_{eff}^{(i)} = M\frac{\hat{\sigma}^2_i}{\hat{C}_i}.
\]

\begin{table}[ht]
\centering
\begin{tabular}{lrrlrr}
  \hline
Outcome & BwTNLV & BwTNML & Outcome & BwTNLV & BwTNML \\ 
  \hline
1 & 73.00 & 509.43 & 9 & 83.47 & 394.90 \\ 
  2 & 65.05 & 472.51 & 10 & 95.39 & 327.35 \\ 
  3 & 87.10 & 1229.39 & 11 & 70.74 & 356.56 \\ 
  4 & 72.64 & 577.73 & 12 & 81.40 & 338.30 \\ 
  5 & 71.67 & 452.01 & 13 & 105.98 & 373.59 \\ 
  6 & 66.49 & 389.78 & 14 & 132.61 & 306.91 \\ 
  7 & 84.30 & 660.37 & 15 & 82.15 & 365.30 \\ 
  8 & 75.46 & 515.09 & 16 & 98.05 & 304.57 \\ 
   \hline
\end{tabular}
\caption{The effective sample sizes for a chain of length $M = 1000$ regarding all $16$ probabilities for possible vector outcomes of visibles. BwTNLV would require at least $4.7$ times as many MCMC iterations to achieve the same amount of effective information about the posterior distribution.} 
\label{tab:m-eff}
\end{table}

For both BwTNLV and BwTNML methods, effective sample sizes for a chain
of length \(M = 1000\) for inference about each of the \(2^4 = 16\)
model probabilities are presented in Table \ref{tab:m-eff}. These range
from \(304.57\) to \(1229.39\) for BwTNML, while BwTNLV only yields
between \(65.05\) and \(132.61\) effective draws. Thus, BwTNLV would
require at least \(4.7\) times as many iterations of the MCMC chain to
be run in order to achieve the same amount of effective information
about the posterior distribution. For this reason, consistent with the
ACF results in Figure \ref{fig:acf}, BwTNLV does not seem to be an
effective method for fitting the RBM, though computing resources can
hinder use of the alternative BwTNML involving marginalization of hidden
variables.

\begin{figure}
\centering
\includegraphics{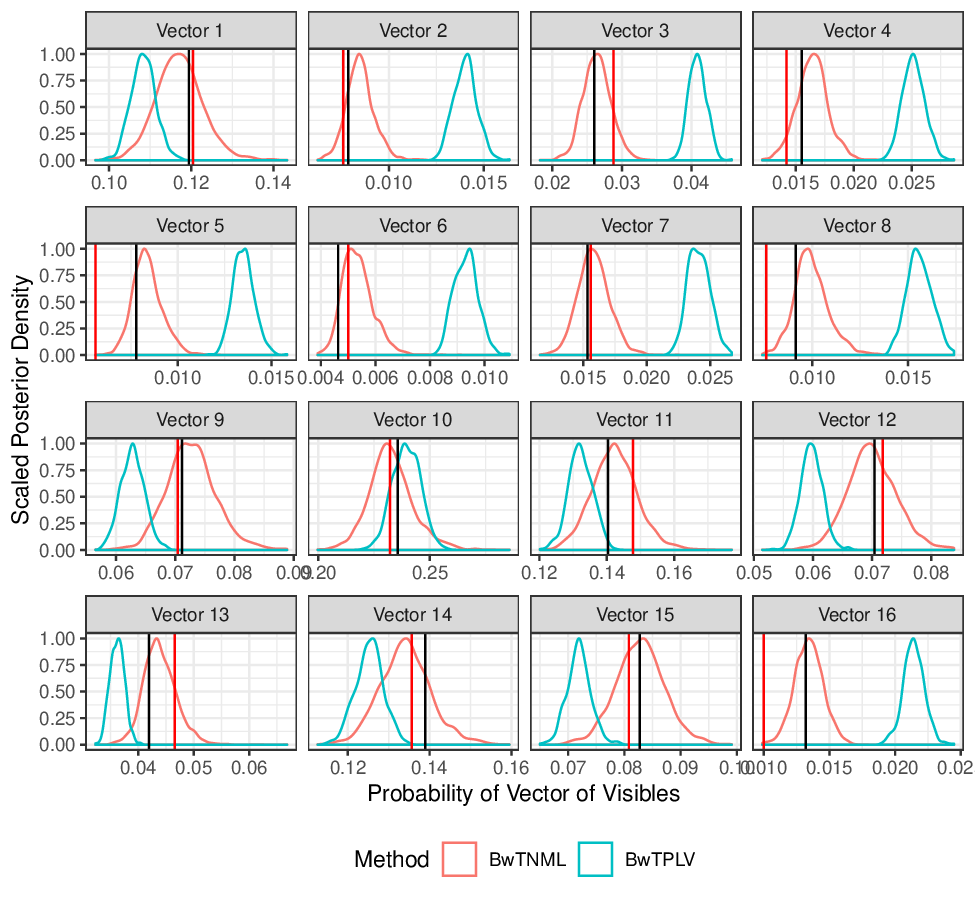}
\caption{\label{fig:fitting-plot}Posterior probabilities of \(16 = 2^4\)
possible realizations of \(4\) visibles using two of the three Bayesian
fitting techniques, BwTPLV and BwTNML. The black vertical lines show the
true probabilities of each vector of visibles based on the parameters
used to generate the training data while the red vertical lines show the
empirical distribution. BwTNML produces the best fit for the data,
however is also the most computationally intensive and least feasible
with a real dataset.}
\end{figure}

Figure \ref{fig:fitting-plot} shows the posterior probability of each
possible \(\boldsymbol v \in \{-1,1\}^4\) after fitting the RBM model
according to method 1 (BwTPLV using trick prior) and method 3 (BwTNML)
(excluding method 2 (BwTNLV) that seeks the same posterior as method 3)
. The black vertical lines show the true probabilities of each image
based on the parameters used to generate the training set while the red
vertical lines show the empirical distribution for the training set of
\(5,000\) vectors. From these posterior predictive checks, it is evident
that BwTNML produces the best fit to the data. Furthermore, along with
the discussion of Section \ref{model-fitting}, Figure
\ref{fig:fitting-plot} also shows that it can be undesirable to seek to
perfectly re-create an empirical distribution in fitting RBM models
(i.e., true model probabilities may differ substantially). The priors in
the BwTNML method constrain the RBM model fit to avoid replication of
the empirical distribution and better estimate the underlying true data
generating probabilities. However, this method requires a
marginalization step to obtain the probability function of visible
observations alone, which is unfeasible for a model with \(\nh\) of any
real size.

\par

An alternative possibility to using MCMC to approximate the posterior
for Bayesian inference in a RBM model is to use variational methods for
approximation of the posterior (Salimans, Kingma, and Welling 2015). The
advantage to using variational methods would be computational
scalability, but the disadvantages are potentially numerous. Namely, the
lack of theoretical guarantees of convergence to a global optimum (Blei,
Kucukelbir, and McAuliffe 2017) and the difficulty of assessing how
closely approximated the posterior is (Yao et al. 2018) are of concern.
These two issues show potential difficulties in using variational Bayes
methods for interval estimation at present.

\hypertarget{discussion}{%
\section{Discussion}\label{discussion}}

RBM models constitute an interesting class of undirected graphical
models that are thought to be useful for supervised learning tasks.
However, when viewed as generative statistical models, RBMs can be
susceptible to forms of model impropriety such as near-degeneracy,
S-instability, and uninterpretability. These model instability problems
relate to how useful the model may be for representing realistic data
generation mechanisms. In this paper, we have provided a theoretical
framework for discussing the model properties of RBMs, as well as
provided empirical evidence that these model problems can arise away
from the origin of the parameter space.

Additionally, these models are difficult to fit using a rigorous
methodology that can incorporate estimation uncertainty, due to the
dimension of the parameter space coupled with the size of the latent
variable space. We have presented three fully Bayes-principled
MCMC-based methods for fitting RBMs, which provide posterior
distributions for quantifying uncertainty in estimating model parameters
and probabilities (e.g., Figure \ref{fig:fitting-plot}) while aiming to
avoid areas of the parameter space that will potentially lead to model
impropriety. We have shown small examples where this fitting methodology
does work, but becomes quickly intractable in large data cases.

The current common practice is to use a kind of MCMC to overcome fitting
complexities, and due to the extreme flexibility in this model class,
rigorous likelihood-based fitting for a RBM can typically seek to merely
merely return the (discrete) empirical distribution for visibles.
Current fitting methodology for RBMs does not provide uncertainty
quantification for the fitted parameters, and it is unclear that they
seeks to address the issues in model impropriety detailed in this paper.
Practitioners should be aware of this and employ some form of
regularization or penalization in the fitting.

Ultimately, it is not unquestionably clear that RBM models are useful as
generative models. As a concern, without appropriate generative behavior
in a RBM model, the uncertainty in estimated model parameters becomes
impossible to realistically assess and the model can further lose useful
application to prediction problems (e.g., realistic predictive
distributions). In the case of classification, predictive distributions
may not be the ultimate goal, but when S-instability is present, small
(imperceptible) differences in the data may lead to greatly different
probabilities and thus greatly different classifications. For these
reasons, we are skeptical about RBMs as probabilistic conceptualizations
for data generation.

\hypertarget{references}{%
\section*{References}\label{references}}
\addcontentsline{toc}{section}{References}

\hypertarget{refs}{}
\leavevmode\hypertarget{ref-blei2017variational}{}%
Blei, David M, Alp Kucukelbir, and Jon D McAuliffe. 2017. ``Variational
Inference: A Review for Statisticians.'' \emph{Journal of the American
Statistical Association} 112 (518). Taylor \& Francis: 859--77.

\leavevmode\hypertarget{ref-carreira2005contrastive}{}%
Carreira-Perpinan, Miguel A, and Geoffrey E Hinton. 2005. ``On
Contrastive Divergence Learning.'' In \emph{Aistats}, 10:33--40.

\leavevmode\hypertarget{ref-cho2012tikhonov}{}%
Cho, KyungHyun, Alexander Ilin, and Tapani Raiko. 2012. ``Tikhonov-Type
Regularization for Restricted Boltzmann Machines.'' \emph{Artificial
Neural Networks and Machine Learning--ICANN 2012}. Springer, 81--88.

\leavevmode\hypertarget{ref-fisher1922mathematical}{}%
Fisher, R. A. 1922. ``On the Mathematical Foundations of Theoretical
Statistics.'' \emph{Philosophical Transactions of the Royal Society of
London A: Mathematical, Physical and Engineering Sciences} 222
(594-604). The Royal Society: 309--68.

\leavevmode\hypertarget{ref-gelman2011inference}{}%
Gelman, Andrew, Kenneth Shirley, and others. 2011. ``Inference from
Simulations and Monitoring Convergence.'' \emph{Handbook of Markov Chain
Monte Carlo}, 163--74.

\leavevmode\hypertarget{ref-box1967discrimination}{}%
G. E. P. Box, W. J. Hill. 1967. ``Discrimination Among Mechanistic
Models.'' \emph{Technometrics} 9 (1). {[}Taylor \& Francis, Ltd.,
American Statistical Association, American Society for Quality{]}:
57--71.

\leavevmode\hypertarget{ref-handcock2003assessing}{}%
Handcock, Mark S. 2003. ``Assessing Degeneracy in Statistical Models of
Social Networks.'' Center for Statistics; the Social Sciences,
University of Washington. \url{http://www.csss.washington.edu/}.

\leavevmode\hypertarget{ref-hinton2010practical}{}%
Hinton, Geoffrey. 2010. ``A Practical Guide to Training Restricted
Boltzmann Machines.'' \emph{Momentum} 9 (1): 926.

\leavevmode\hypertarget{ref-hinton2002training}{}%
Hinton, Geoffrey E. 2002. ``Training Products of Experts by Minimizing
Contrastive Divergence.'' \emph{Neural Computation} 14 (8). MIT Press:
1771--1800.

\leavevmode\hypertarget{ref-hinton2006fast}{}%
Hinton, Geoffrey E, Simon Osindero, and Yee-Whye Teh. 2006. ``A Fast
Learning Algorithm for Deep Belief Nets.'' \emph{Neural Computation} 18
(7). MIT Press: 1527--54.

\leavevmode\hypertarget{ref-kaiser2007statistical}{}%
Kaiser, Mark S. 2007. ``Statistical Dependence in Markov Random Field
Models.'' \emph{Statistics Preprints} Paper 57. Digital Repository @
Iowa State University.
\url{http://lib.dr.iastate.edu/stat_las_preprints/57/}.

\leavevmode\hypertarget{ref-kaplan2016propriety}{}%
Kaplan, Andee, Daniel Nordman, and Stephen Vardeman. 2017. ``On the
Propriety of Restricted Boltzmann Machines.'' \emph{Unpublished}.

\leavevmode\hypertarget{ref-larochelle2008classification}{}%
Larochelle, Hugo, and Yoshua Bengio. 2008. ``Classification Using
Discriminative Restricted Boltzmann Machines.'' In \emph{Proceedings of
the 25th International Conference on Machine Learning}, 536--43. ACM.

\leavevmode\hypertarget{ref-lehmann1990model}{}%
Lehmann, E. L. 1990. ``Model Specification: The Views of Fisher and
Neyman, and Later Developments.'' \emph{Statistical Science} 5 (2).
Institute of Mathematical Statistics: 160--68.

\leavevmode\hypertarget{ref-le2008representational}{}%
Le Roux, Nicolas, and Yoshua Bengio. 2008. ``Representational Power of
Restricted Boltzmann Machines and Deep Belief Networks.'' \emph{Neural
Computation} 20 (6). MIT Press: 1631--49.

\leavevmode\hypertarget{ref-li2014biclustering}{}%
Li, Jing. 2014. ``Biclustering Methods and a Bayesian Approach to
Fitting Boltzmann Machines in Statistical Learning.'' PhD thesis, Iowa
State University; Graduate Theses; Dissertations.
\url{http://lib.dr.iastate.edu/etd/14173/}.

\leavevmode\hypertarget{ref-montufar2011refinements}{}%
Montufar, Guido, and Nihat Ay. 2011. ``Refinements of Universal
Approximation Results for Deep Belief Networks and Restricted Boltzmann
Machines.'' \emph{Neural Computation} 23 (5). MIT Press: 1306--19.

\leavevmode\hypertarget{ref-montufar2011expressive}{}%
Montúfar, Guido F, Johannes Rauh, and Nihat Ay. 2011. ``Expressive Power
and Approximation Errors of Restricted Boltzmann Machines.'' In
\emph{Advances in Neural Information Processing Systems}, 415--23. NIPS.

\leavevmode\hypertarget{ref-nguyen2014deep}{}%
Nguyen, Anh Mai, Jason Yosinski, and Jeff Clune. 2014. ``Deep Neural
Networks Are Easily Fooled: High Confidence Predictions for
Unrecognizable Images.'' \emph{arXiv Preprint arXiv:1412.1897}.
\url{http://arxiv.org/abs/1412.1897}.

\leavevmode\hypertarget{ref-salakhutdinov2012efficient}{}%
Salakhutdinov, Ruslan, and Geoffrey Hinton. 2012. ``An Efficient
Learning Procedure for Deep Boltzmann Machines.'' \emph{Neural
Computation} 24 (8). MIT Press: 1967--2006.

\leavevmode\hypertarget{ref-salakhutdinov2009deep}{}%
Salakhutdinov, Ruslan, and Geoffrey E Hinton. 2009. ``Deep Boltzmann
Machines.'' In \emph{International Conference on Artificial Intelligence
and Statistics}, 448--55. AI \& Statistics.

\leavevmode\hypertarget{ref-salimans2015markov}{}%
Salimans, Tim, Diederik Kingma, and Max Welling. 2015. ``Markov Chain
Monte Carlo and Variational Inference: Bridging the Gap.'' In
\emph{International Conference on Machine Learning}, 1218--26.

\leavevmode\hypertarget{ref-schweinberger2011instability}{}%
Schweinberger, Michael. 2011. ``Instability, Sensitivity, and Degeneracy
of Discrete Exponential Families.'' \emph{Journal of the American
Statistical Association} 106 (496). Taylor \& Francis: 1361--70.

\leavevmode\hypertarget{ref-sculley2018winner}{}%
Sculley, D., Jasper Snoek, Alex Wiltschko, and Ali Rahimi. 2018.
``Winner's Curse? On Pace, Progress, and Empirical Rigor.'' In \emph{6th
International Conference on Learning Representations Workshop
Submission}. \url{https://openreview.net/forum?id=rJWF0Fywf}.

\leavevmode\hypertarget{ref-smolensky1986information}{}%
Smolensky, Paul. 1986. ``Information Processing in Dynamical Systems:
Foundations of Harmony Theory.'' DTIC Document.

\leavevmode\hypertarget{ref-srivastava2012multimodal}{}%
Srivastava, Nitish, and Ruslan R Salakhutdinov. 2012. ``Multimodal
Learning with Deep Boltzmann Machines.'' In \emph{Advances in Neural
Information Processing Systems}, 2222--30. NIPS.

\leavevmode\hypertarget{ref-srivastava2013modeling}{}%
Srivastava, Nitish, Ruslan R Salakhutdinov, and Geoffrey E Hinton. 2013.
``Modeling Documents with Deep Boltzmann Machines.'' \emph{arXiv
Preprint arXiv:1309.6865}. \url{http://arxiv.org/abs/1309.6865}.

\leavevmode\hypertarget{ref-szegedy2013intriguing}{}%
Szegedy, Christian, Wojciech Zaremba, Ilya Sutskever, Joan Bruna,
Dumitru Erhan, Ian J. Goodfellow, and Rob Fergus. 2013. ``Intriguing
Properties of Neural Networks.'' \emph{arXiv Preprint arXiv:1312.6199}.
\url{http://arxiv.org/abs/1312.6199}.

\leavevmode\hypertarget{ref-tieleman2008training}{}%
Tieleman, Tijmen. 2008. ``Training Restricted Boltzmann Machines Using
Approximations to the Likelihood Gradient.'' In \emph{Proceedings of the
25th International Conference on Machine Learning}, 1064--71. ACM.

\leavevmode\hypertarget{ref-pmlr-v80-yao18a}{}%
Yao, Yuling, Aki Vehtari, Daniel Simpson, and Andrew Gelman. 2018.
``Yes, but Did It Work?: Evaluating Variational Inference.'' In
\emph{Proceedings of the 35th International Conference on Machine
Learning}, edited by Jennifer Dy and Andreas Krause, 80:5581--90.
Proceedings of Machine Learning Research. Stockholmsmässan, Stockholm
Sweden: PMLR. \url{http://proceedings.mlr.press/v80/yao18a.html}.

\leavevmode\hypertarget{ref-zhou2014some}{}%
Zhou, Wen. 2014. ``Some Bayesian and Multivariate Analysis Methods in
Statistical Machine Learning and Applications.'' PhD thesis, Iowa State
University; Graduate Theses; Dissertations.
\url{http://lib.dr.iastate.edu/etd/13816/}.

%\bibliography{../resources/latex/refs\_rbm.bib}

\end{document}

%% file: rbm.tikz
\begin{tikzpicture}[auto, node distance=3cm, thick, 
                    main node/.style= {circle,
                      fill=gray!30,
                      draw,
                      font=\sffamily\Large\bfseries,
                      minimum size=1cm}]

  \node[main node] (1) {$h_j$};
  \node (11) [above of=1, yshift = -2cm] {$\theta_{h_j}$};
  \node[main node] (2) [right of=1, xshift = -1.5cm] {};
  \node[main node] (3) [right of=2, xshift = -1.5cm] {};
  \node[main node] (4) [right of=3, xshift = -1.5cm] {};
  \node[main node] (5) [right of=4, xshift = -1.5cm] {};
  \node (0) [right of=5, xshift = -.5cm] {Hidden Layer $\mathcal{H}$};
  \node[main node] (6) [below left of=1, fill=white, xshift = .5cm, yshift = -1cm] {$v_i$};
  \node (12) [below of=6, yshift = 2.25cm] {$\theta_{v_i}$};
  \node[main node] (7) [right of=6, fill=white] {};
  \node[main node] (8) [right of=7, fill=white] {};
  \node[main node] (9) [right of=8, fill=white] {};
  \node (10) [right of=9, xshift = -.5cm] {Visible Layer $\mathcal{V}$};
  
  \path
    (1) edge node [left=.5cm] {$\theta_{ij}$} (6)
        edge node {} (7)
        edge node {} (8)
        edge node {} (9)
    (2) edge node {} (6)
        edge node {} (7) 
        edge node {} (8)
        edge node {} (9)
    (3) edge node {} (6)
        edge node {} (7)
        edge node {} (8)
        edge node {} (9)
    (4) edge node {} (6)
        edge node {} (7) 
        edge node {} (8)
        edge node {} (9)
    (5) edge node {} (6)
        edge node {} (7) 
        edge node {} (8)
        edge node {} (9);
\end{tikzpicture}

%% file: toymodel.tikz
\begin{tikzpicture}[auto, node distance=3cm, thick, 
                    main node/.style= {circle,
                      fill=gray!30,
                      draw,
                      font=\sffamily\Large\bfseries}]

  \node[main node] (1) {$h_1$};
  \node[main node] (2) [below of=1, fill=white] {$v_1$};
  
  \path
    (1) edge node [left=.5cm] {$\theta_{11}$} (2);
\end{tikzpicture}

%% file: toyhull_top.tikz
\begin{tikzpicture}[tdplot_main_coords]
      \draw[dotted,->,black] (0,0,0) -- (3,0,0) node[anchor=west]{$x$};
      \draw[dotted,->] (0,0,0) -- (0,3,0) node[anchor=north east]{$y$};
      \draw[dotted,->] (0,0,0) -- (0,0,3) node[anchor=east]{$z$};
      
      \draw[thin,color=lightgray] (0,0,0) -- (1,0,0) -- (1,1,0) -- (0,1,0) -- (0,0,0);
      \draw[thin,color=lightgray] (0,0,0) -- (0,0,1) -- (1,0,1) -- (1,0,0) -- (0,0,0);
      \draw[thin,color=lightgray] (0,0,0) -- (0,0,1) -- (0,1,1) -- (0,1,0) -- (0,0,0);
      \draw[thin,color=lightgray] (0,0,1) -- (1,0,1) -- (1,1,1) -- (0,1,1) -- (0,0,1);
      \draw[thin,color=lightgray] (0,1,0) -- (1,1,0) -- (1,1,1) -- (0,1,1) -- (0,1,0);
      \draw[thin,color=lightgray] (1,1,0) -- (1,1,1) -- (1,0,1) -- (1,0,0) -- (1,1,0);